# Evaluating Temporal Graphs Built from Texts via Transitive Reduction

**Xavier Tannier** XTANNIER@LIMSI.FR
*LIMSI-CNRS and Univ. Paris-Sud*
*B.P. 133*
*91403 ORSAY Cedex, France*

**Philippe Muller** MULLER@IRIT.FR
*ALPAGE-INRIA and Toulouse University*
*IRIT, Univ. Paul Sabatier*
*118 Route de Narbonne*
*F-31062 Toulouse Cedex 04, France*

## Abstract

Temporal information has been the focus of recent attention in information extraction, leading to some standardization effort, in particular for the task of relating events in a text. This task raises the problem of comparing two annotations of a given text, because relations between events in a story are intrinsically interdependent and cannot be evaluated separately. A proper evaluation measure is also crucial in the context of a machine learning approach to the problem. Finding a common comparison referent at the text level is not obvious, and we argue here in favor of a shift from event-based measures to measures on a unique textual object, a minimal underlying temporal graph, or more formally the transitive reduction of the graph of relations between event boundaries. We support it by an investigation of its properties on synthetic data and on a well-know temporal corpus.

## 1. Introduction

Temporal processing of texts is a somewhat recent field from a methodological point of view, even though temporal semantics has a long tradition, dating back at least to the 1940's (Reichenbach, 1947). While theoretical and formal linguistic approaches to temporal interpretation at the discourse level have been very active in the late 1980s and early 1990s (Kamp & Reyle, 1993; Asher & Lascarides, 1993; Steedman, 1995; Webber, 1988), empirical approaches were less frequent, and very few natural language processing systems were evaluated beyond a few instances (Grover, Hitzeman, & Moens, 1995; Kameyama, Passonneau, & Poesio, 1993; Passonneau, 1988; Song & Cohen, 1991).

Temporal information being essential to the interpretation of a text and thus crucial in applications such as summarization or information extraction, it has received growing attention in the 2000s (Mani, Pustejovsky, & Gaizauskas, 2005) and has lead to some standardization effort through the TimeML initiative (Saurí, Littman, Knippen, Gaizauskas, Setzer, & Pustejovsky, 2006). We address here a central part in this task, namely evaluating the extraction of the network of temporal relations between events described in a text. Since temporal information is not easily broken down into local bits of information, there are many equivalent ways to express the same ordering of events. Human annotation is thus notoriously difficult (Setzer, Gaizauskas, & Hepple, 2006) and comparisons between annotations cannot rely on simple precision/recall-type measures. The given practice nowadays has been to compute some sort of transitive closure over the network/graph of constraints on temporal events (usually expressed in the well-known Allen algebra (Allen, 1983), or a sub-algebra), and then either to compare the sets of simple temporal relations that are deduced from it with standard precision and recall, or to measure the agreement between all relations, including disjunctions of information (Verhagen, Gaizauskas, Schilder, Hepple, Katz, & Pustejovsky, 2007). This reasoning model is also used to help build representations of temporal situations by imposing global constraints on top of local decision problems (Chambers & Jurafsky, 2008a; Tatu & Srikanth, 2008; Bramsen, Deshpande, Lee, & Barzilay, 2006).

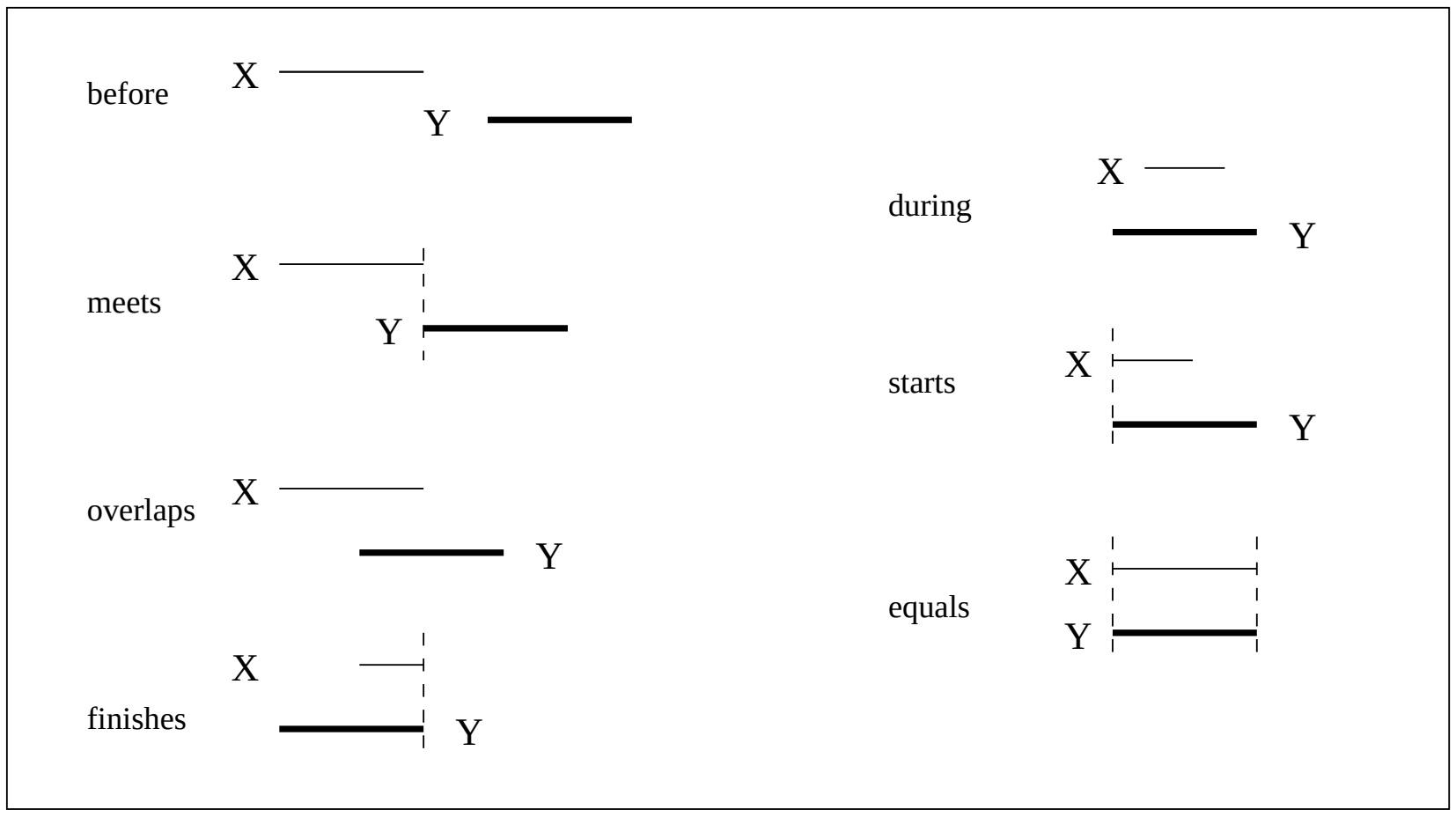

Figure 1: Allen relations. Each relation $r$ has an inverse relation $ri$.

We take a different route here, by extracting a single referent graph, a *minimal graph* of constraints. There are a number of ways of doing this and we argue for basing it on the graph of relations between event boundaries. We aim to accomplish two things by doing so: to find a graph that is easy to compute, and to eliminate a bias introduced by measures that do not take into account the combinatorial aspect of agreement on transitive closure graphs.

The next section presents in more detail the usual way of comparing annotation graphs between temporal entities extracted from a text, and the problems it raises. Then we argue for comparing event boundaries instead of events and define two new metrics that apply to that type of information. We focus on convex relations, a tractable sub-algebra of Allen relations which covers human annotations. Finally, we present an empirical study of the behavior of these measures on generated data and on the TimeBank Corpus (Pustejovsky, Hanks, Saurí, See, Gaizauskas, Setzer, Radev, Sundheim, Day, Ferro, & Lazo, 2003) to support our claim of the practicality of this methodology.

## 2. Comparing Temporal Constraint Networks

Work on temporal annotation of texts strongly relies on Allen's interval algebra. Allen represents time and events as intervals, and states that 13 basic relations can hold between these intervals (see Figure 1 and Table 1), by considering every possible ordering of the interval endpoints. These binary relations, existing amongst all intervals of a collection (in our case, corresponding to temporal entities in a text), define a graph where nodes are the intervals and where edges are labelled with the set of relations which may hold between a pair of nodes. The relations are mutually exclusive. The TimeML specification for linguistic temporal annotation uses Allen relations under different names, and other projects either use a subset or groupings of these relations (see below).

We are interested in this paper in evaluating systems annotating texts by temporal relations holding between events or between temporal expressions and events. For example, consider the following text, extracted from the TimeBank corpus:

| Relation | Meaning | Endpoint relations | Inverse relation |
|---|---|---|---|
| $I\ b\ J$ | I before J | $I_2 < J_1$ | $bi$ |
| $I\ m\ J$ | I meets J | $I_2 = J_1$ | $mi$ |
| $I\ o\ J$ | I overlaps J | $(I_1 < J_1) \wedge (I_2 < J_2) \wedge (J_1 < I_2)$ | $oi$ |
| $I\ s\ J$ | I starts J | $(I_1 = J_1) \wedge (I_2 < J_2)$ | $si$ |
| $I\ d\ J$ | I during J | $(J_1 < I_1) \wedge (I_2 < J_2)$ | $di$ |
| $I\ f\ J$ | I finishes J | $(J_1 < I_1) \wedge (I_2 = J_2)$ | $fi$ |
| $I\ =\ J$ | I equals J | $(I_1 = J_1) \wedge (I_2 = J_2)$ | |

Table 1: Allen relations. Each relation $r$ has an inverse relation $ri$. An interval $I$ starts at $I_1$ and ends at $I_2$.

(1) It wasn't until twenty years after the first astronauts were **chosen**$_{e_1}$ that NASA finally **included**$_{e_2}$ six women, and they were all scientists, not pilots. No woman has actually been **in charge of a mission**$_{e_3}$ until **now**$_{t_1}$.

A correct annotation of temporal relations could be given by the graph shown in Figure 2. Other relations could be explicited, i.e. $e_1 b t_1$, and a complete evaluation could consider all possible edges.

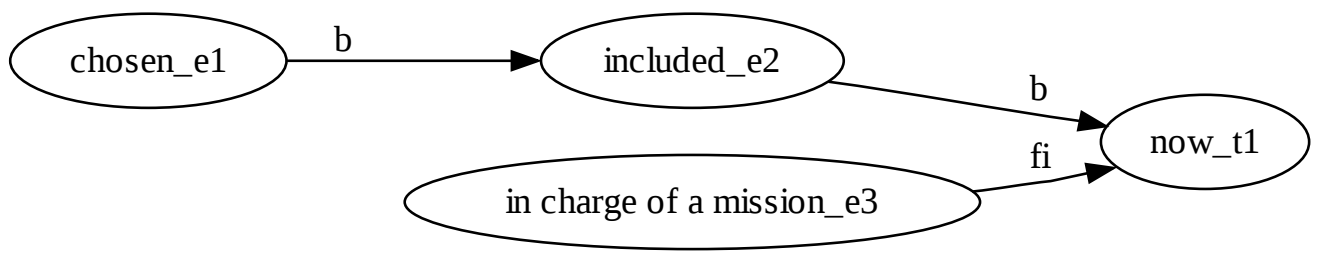

Figure 2: Example annotation for example 1.

Precision and recall evaluations are often not performed on graphs of relations between all events in a text however, but on the subproblem of ordering pairs of successively described events (Mani & Schiffman, 2005; Verhagen et al., 2007) or even same-sentence events (Lapata & Lascarides, 2006) (in our example, only $e_1\ b\ e_2$ and $e_3\ fi\ t_1$ [1]). The main reason of this choice is the difficulty of the task, even for human beings, of assigning temporal relations in a large text (Setzer et al., 2006). Another issue is that evaluation of full temporal graphs is an open question. As will be further discussed in this section, metrics traditionally used for the task, namely recall and precision metrics (either strict or relaxed), raise specific problems that are still to be addressed.

We detail now important notions concerning temporal networks and the comparison of these networks. All example relations given in this section are expressed in terms of Allen algebra, whose set of relations and their abbreviations are recalled in Table 1. Here, we use the classical symbols $<$ and $>$ for order on temporal points.

## 2.1 Temporal Closure

Temporal closure is an inferential closure mechanism that consists in composing known pairs of temporal relations in order to obtain new relations, up to a fixed point. *E.g.*: if $A\,b\,B$ and $C\,d\,B$, then

1. Exceptions exist, as in the work of Mani et al. (2006) and Mani et al. (2007).

| $\nearrow$ | b | bi | d | di |
|---|---|---|---|---|
| b | b | all | {b, o, m, d, s } | b |
| bi | all | bi | {bi, oi, mi, d, f} | bi |
| d | b | bi | d | all |
| di | {b, o, m, di, fi} | {bi, oi, di, mi, si} | {o, oi, d, s, f, di, si, fi, =} | di |

Table 2: Composition between a few Allen relations.

$A\,b\,C$; the operation can lead to a disjunction of relations, for example if $A\,b\,B$ and $B\,d\,C$ then $A\,b\,C \vee A\,o\,C \vee A\,m\,C \vee A\,d\,C \vee A\,s\,C$. This can also be noted as $A\{b, o, m, d, s\}C$.

If we consider *generalized relations*, *i.e.* the set $\mathcal{R}$ of disjunctions of basic temporal relations, each seen as a set of base relations, then set union and intersection and the composition of relations define an algebra on $\mathcal{R}$ (the algebra of the subsets of the set of all Allen relations). Composition of relations is the operation that generalizes inferences from basic relations to sets of relations. Taking the previous example, if now $A\,T\,B$ and $B\,S\,C$, with $T = \{t_1, t_2, ...t_k\}$ and $S = \{s_1, s_2, ...s_m\}$, $t_i$, $s_i$ being base relations:

$$T \circ S = \{t_1, t_2, ...t_k\} \circ \{s_1, s_2, ...s_m\} = \bigcup_{i,j} (t_i \circ s_j)$$

So any composition of relations can be computed from the 13x13 compositions of base relations.

A table of all composition rules in Allen algebra can be found in the work of Allen (1983) or Rodriguez et al. (2004), and a sample for a few basic relations is given in Table 2. These new relations do not express new intrinsic constraints, but make the temporal situation more explicit. They can also make some information more precise, as disjunctions inferred about the same edge are intersected to combine inferences from different compositions.

A constraint propagation algorithm ensures that all existing temporal relations are added to the network, labelling an inconsistency with $\emptyset$ (Allen, 1983). This path-consistency algorithm is sound, but not complete, as it does not detect all cases of inconsistency. See the simple version presented in Algorithm 1. More efficient versions have also been developed (Vilain, Kautz, & van Beek, 1990), and will be put to use in our experiments, but it is not our main focus here.

It is not desirable to compare temporal graphs without performing a temporal closure on them. Indeed, there are several ways to encode the same temporal information in a graph, as shown in the (very simple) example of Figure 3. Closure can be seen as a computationally simple way of expliciting the temporal information from an annotation, allowing for more precise comparisons. But temporal closure also produces redundant information, which can lead to evaluation issues, as will be explained in Section 2.4.

In this paper, we call $G^*$ the temporal closure (also called *a saturated graph*) of a graph $G$.

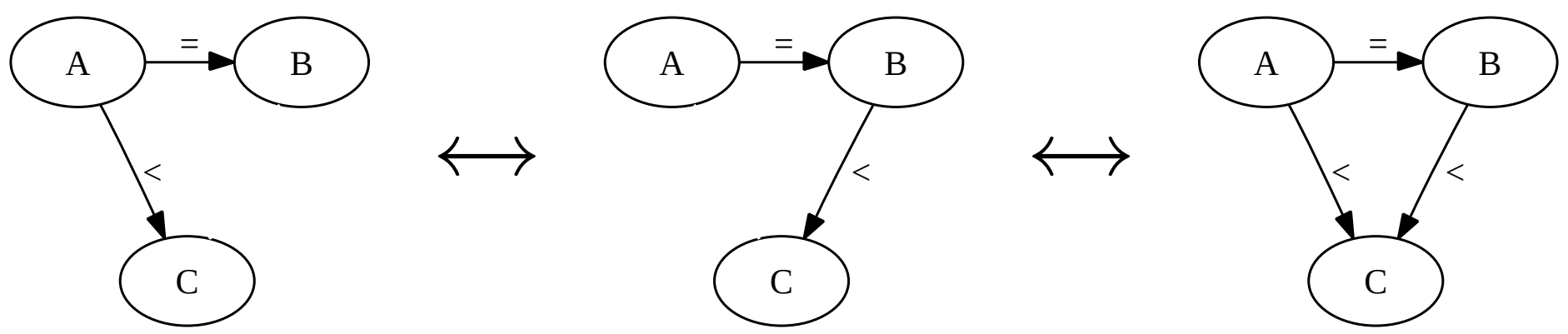

Figure 3: Three identical annotations. The last one is the result of temporal closure.

**Algorithm 1** Temporal closure

Let $U$ = the disjunction of all 13 Allen relations,
$R_{m,n}$ = the current relation between nodes $m$ and $n$

```
procedure CLOSURE(G)
    A=G.edges()
    N=G.vertices()
    changed = True
    while changed do
        changed = False
        for all pairs of nodes (i, j) ∈ N × N do
            for all k ∈ N such that ((i, k) ∈ A ∧ (k, j) ∈ A) do        ▷ composition via k
                R1_{i,j} = (R_{i,k} ∘ R_{k,j})                          ▷ to find info about (i,j)
                if no edge (a relation R2_{i,j}) existed before between i and j then R2_{i,j} = U
                end if
                R_{i,j} = R1_{i,j} ∩ R2_{i,j}                            ▷ intersect new with already known
                if R_{i,j} = ∅ then error                                 ▷ inconsistency detected
                else if R_{i,j} = U then do nothing                      ▷ no new information
                else
                    update edge (i,j)
                    changed = True
                end if
            end for
        end for
    end while
end procedure
```

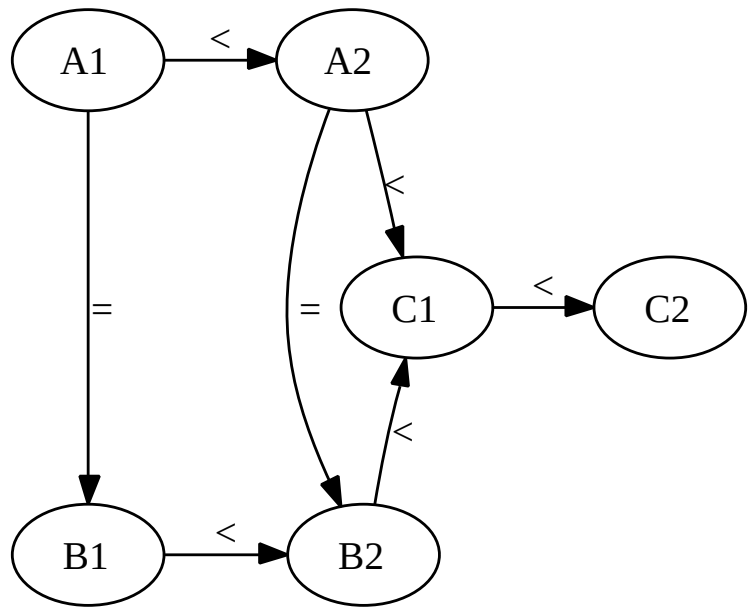

Figure 4: Endpoint graph (same temporal information as Figure 3).

## 2.2 Time Point Algebra and Convex Relations

Interval graphs can be converted easily into graphs between points (Vilain et al., 1990), where an event is split into a beginning and an ending point; the mapping between Allen relations and point relations is given at Table 1. This leads to a smaller set of simple relations: equality ($=$) and precedence ($<$ and $>$), and a simpler algebra, with only 7 consistent sets ($\{<\}$,$\{<,=\}$,$\{=\}$,$\{<,=,>\}$,$\{>\}$,$\{>=\}$, $\{<,>\}$ where a set denotes a disjunction of relations). The point algebra is obtained by the four relations $r_1$ ... $r_4$ that hold between the endpoints of the two intervals $I$ and $J$ started by $I_b$ and $J_b$ and ended by $I_e$ and $J_e$ respectively. These four relations are $I_b\ r_1\ J_b$, $I_e\ r_2\ J_e$, $I_b\ r_3\ J_e$ and $I_e\ r_4\ J_b$. Converting an Allen graph into an endpoint graph is thus straightforward. Figure 4 shows a point graph equivalent to the interval graph of Figure 3.

As with the interval algebra, pairs of point relations can be combined and the temporal closure can be computed in the same way. The relation between two time points is *continuous* if the assigned set of simple relations is convex (Vilain et al., 1990).

A so-called *convex relation* corresponds to cases where relations $r_1$ to $r_4$ are assigned to one of the 6 possible relations $\{<\}, \{<,=\}, \{=\}, \{<,=,>\}, \{>\}, \{>=\}$[2], considered as *conceptual neighbors* (Freksa, 1992). Using only these relations between endpoints restricts interval relations to sets of Allen relations that are *conceptual neighbors*. This means that they encode relations that may be vague but in which intervals endpoints can only be in convex subsets of the time-line. Figure 5 shows which Allen relations are conceptual neighbors. Another useful way of seeing these conceptual neighbors is by considering continuous transformations of an interval endpoints on the time-line: when a relation $r$ holds between two intervals $I_1$ and $I_2$, moving continuously their endpoints can only change the relation to a conceptual neighbor of $r$. For instance, such a conceptual transformation cannot change a situation where $I_1$ starts $I_2$ to a situation where $I_1 < I_2$ without going through (at least) intermediary situations $I_1$ overlaps $I_2$ and $I_1$ meets $I_2$.

Finally, instead of $2^{13}$ possible disjunctive relations in Allen algebra, the set of corresponding interval convex relations is reduced to 82. The corresponding sub-algebra is tractable, that is the problem of the satisfiability of a set of constraints has a sound and complete polynomial time algorithm[3]. Moreover, it will ensure the uniqueness of our minimal graphs, that will be defined and described in this paper.

It is important to note that a temporal graph built from an annotated text contains only convex relations, since the graph is generated from a finite set of base relations (annotators are not allowed to

2. The 7 relations described above, except $\{<,>\}$, also noted $\neq$. These 6 relations form a sub-algebra: all compositions of disjunctions of these relations are disjunctions of these relations.
3. See the work by Schilder (1997) for a more complete presentation within a natural language processing perspective.

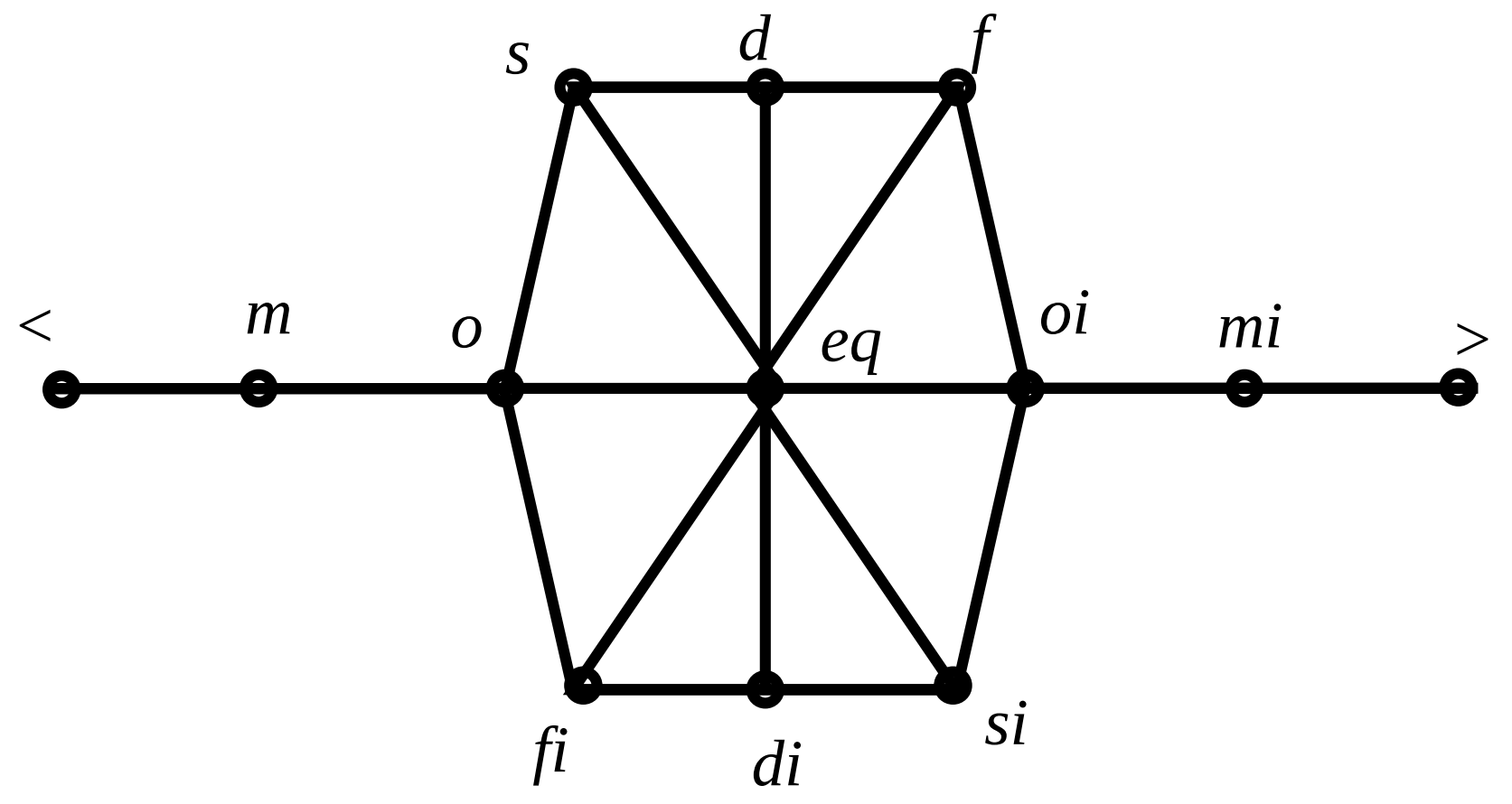

Figure 5: Temporal relations which are conceptual neighbors

use disjunctions), and since the set of all convex relations is stable under composition and thus forms a sub-algebra.

### 2.3 Strict and Relaxed Recall and Precision

In the general case, both humans and systems may assign disjunctions of atomic relations between two events (*i.e.* $A\,b\,B\ \vee\ A\,m\,B$), directly or indirectly (after saturation). This is a way to reduce vagueness even if the exact relation is not known.

The presence of disjunctions raises the question of how to score relations that are only partly correct, like $A\,b\,B\ \vee\ A\,m\,B$ instead of $A\,b\,B$ or the reverse. In response to that issue, different variations of usual precision and recall measures have been proposed.

A *strict* measure only counts exact matching as success, and will for example score 0 for the latter example.

However, it can be argued that an evaluation measure should take better account of "close matches". For example, suppose that the gold standard relation between $A$ and $B$ is $A\,b\,B$. If the system chooses the disjunction $A\,b\,B\ \vee\ A\,m\,B$, it must be rewarded less than $A\,b\,B$ but more than $A > B$ or nothing. The system is vaguer but correct, as its annotation is a logical consequence of the standard annotation.

We proposed (Muller & Tannier, 2004) such a gradual measure that we might call "temporal" precision and recall. If $S_{i,j}$ is the (possibly disjunctive) relation between $i$ and $j$ given by the system and $K_{i,j}$ the (possibly disjunctive) gold standard, then:

$$P_{temp\ i,j} = \frac{Card(\,S_{i,j} \cap K_{i,j}\,)}{Card(\,K_{i,j}\,)} \qquad R_{temp\ i,j} = \frac{Card(\,S_{i,j} \cap K_{i,j}\,)}{Card(\,S_{i,j}\,)}$$

With $Card(G_{i,j})$ = the number of atomic relations present in the disjunction. Thus, in our example, the system $S_1$ that answered *before* $\vee$ *overlaps* between $i$ and $j$ will get:

$$P_{temp\ i,j,S_1} = \frac{Card(\{b,o\}\ \cap\ \{b\}\,)}{Card(\ \{b\}\,)} = 1$$

$$R_{temp\ i,j,S_1} = \frac{Card(\{b,o\}\ \cap\ \{b\}\,)}{Card(\{b,o\})} = \frac{1}{2}$$

While $S_2$ that answered *after* will get $P_{temp\ i,j,S_2} = R_{temp\ i,j,S_2} = 0$.

The final precision (resp. recall) is the average on the number of relations given by the system (resp. by the reference):

$$P_{temp} = \frac{\sum_{i=1}^{i=n} \sum_{j=i+1}^{j=n} P_{temp\ i,j}}{Card(S)} \qquad R_{temp} = \frac{\sum_{i=1}^{i=n} \sum_{j=i+1}^{j=n} R_{temp\ i,j}}{Card(K)}$$

Similar measures have also been used during the TempEval evaluation campaign in 2007 (Verhagen et al., 2007), with a reduced set of relations : "before", "overlaps" and "before or overlaps". These measures were called *relaxed* recall and precision. We will use the words *strict* and *relaxed* to designate these two ways to score temporal relations.

## 2.4 Relative Importance of Relations

As shown above, temporal closure is necessary in order to be able to compare properly two temporal graphs. But in a temporal graph, relations do not have all the same importance. Applying basic recall and precision scores (either strict or relaxed) on closed temporal graphs is not enough. Consider the very simple graph examples of Figure 6, in which the first graph $K$ is the gold standard. $S_1$ contains only two relations, against six in $K$. But it seems unfair to consider a recall score of $\frac{2}{6}$, since adding only one relation ($B\, b\, C$) would be enough to infer all others. An intuitive recall would be around $\frac{2}{3}$. What counts here is how many relations are missing in order to recover the whole annotation graph.

Still, even if we suppose that we have a way to distinguish unambiguously "major" relations (solid lines in $K$) from "minor" (deducible) relations (dashed lines), it would not be enough. Indeed, graph $S_2$ finds the relation "$B\, b\, D$". This relation is minor in $K$, because it can be found by composing other relations; but in $S_2$, it is not the case, this relation actually carries a piece of information and must then be rewarded. However, even if the amount of temporal information brought by $S_2$ and $S_3$ seems equivalent, $S_3$ should get a higher score. Indeed, the amount of *missing* relations (needed to infer the full graph) is much lower in $S_3$ (only "$C\, b\, D$" is missing) than in $S_2$. Finally, $S_4$ should get a better recall than any other one. General cases involving all relations are obviously much more complex.

The same kind of problems could be found in the co-reference task of MUC-6, where co-reference links define an equivalence relation. It is thus not necessary to specify all pair-wise co-reference relations to retrieve them, and this has consequences for the evaluation of recall. This was addressed by considering that a spanning tree of the graph of co-reference is enough to evaluate the recall of such links (Vilain, Burger, Aberdeen, Connolly, & Hirschman, 1995). For each equivalence class considered, consisting of $n$ entities, $n-1$ links are enough to define the class, and recall error depends on the number of missing links $m$ needed to reconnect an equivalence class. Recall on a class is then $1 - \frac{m}{n-1}$. Precision was then defined symmetrically. In a much simpler framework, this was a similar problem of redundancy.[4]

## 2.5 "Minimal" Graphs

As said in previous section, a good, but insufficient way to deal with the relative importance of graph relations would be to work on what we called "major" relations, or "minimal graph".

We consider that a temporal graph is a "minimal graph" of a graph $G$ if:

4. The MUC measures has other problems that were later addressed by the measures $B^3$ and CEAF, but which have no relevance to the evaluation of a temporal graph. The main issue is a tendency to favor the prediction of a co-reference link for every pair of mentions; this has no counterpart in the temporal case since event pairs can be linked with different relations and with different inferential properties, as opposed to only one equivalence relation.

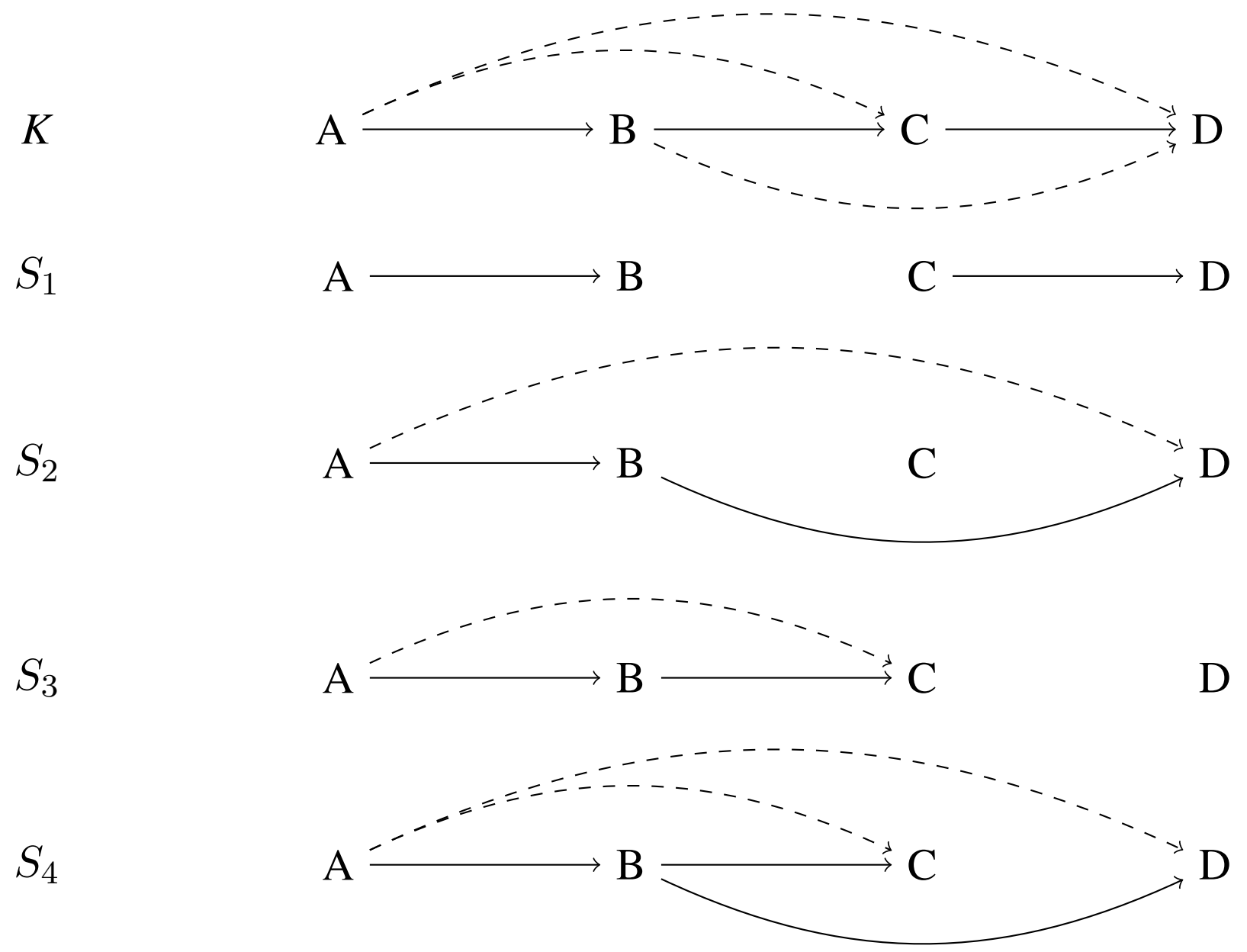

Figure 6: What a metric should deal with. K is the reference annotations, $S_i$ are candidate annotations. $S_4$ is better than $S_1$ and $S_3$, which are better than $S_2$. Solid lines indicate annotated relations, dashed lines indicate relations inferred from the annotated relations. All events are related with the “before” relation here.

1. Its temporal closure leads to the same temporal information as $G$.
2. No relation can be removed from this graph without breaking the first property.

Unfortunately, a unique minimal graph does not exist in the general case, and in particular for Allen relations. Rodriguez et al. (2004) propose a way to find all minimal graphs for a given temporal graph. Their algorithm first finds *core* relations, relations that are in every minimal graph, by intersecting all derivations, and then computes all possible remaining combinations in order to find those composing a minimal graph.

For example, for the relation $R_{A,B}$ between $A$ and $B$, derivations are $R_{A,C} \circ R_{C,B}$, $R_{A,D} \circ R_{D,B}$, $R_{A,E} \circ R_{E,B}$, etc. If the intersection of all these derived relations equals $R_{A,B}$, it means that $R_{A,B}$ is not a *core* relation, since it can be obtained by composing some other relations. Otherwise, the relation is a *core* relation, since removing it always leads to a loss of information. The way this “kernel” is obtained ensures its uniqueness.

However, the second part of the procedure (compute remaining combinations) is computationally impractical, even for medium-sized graphs, since every subset of relations must be considered to determine a minimal graph on top of core relations. The authors do not detail much their empirical investigations, offering no support for the usability of this method. Moreover, it does not lead to a unique graph that could be compared to a reference.

Going back to evaluation, Tannier and Muller (2008) suggest the comparison of graphs through *core* relations, which are easy to compute and give a good idea of how important the relations are in a same graph. But core relations do not contain all the information provided by closed graphs, and measures on core graphs are only an approximation of what should be assessed. In this paper, we propose a method to obtain a unique graph respecting the constraints mentioned at Section 2.4.

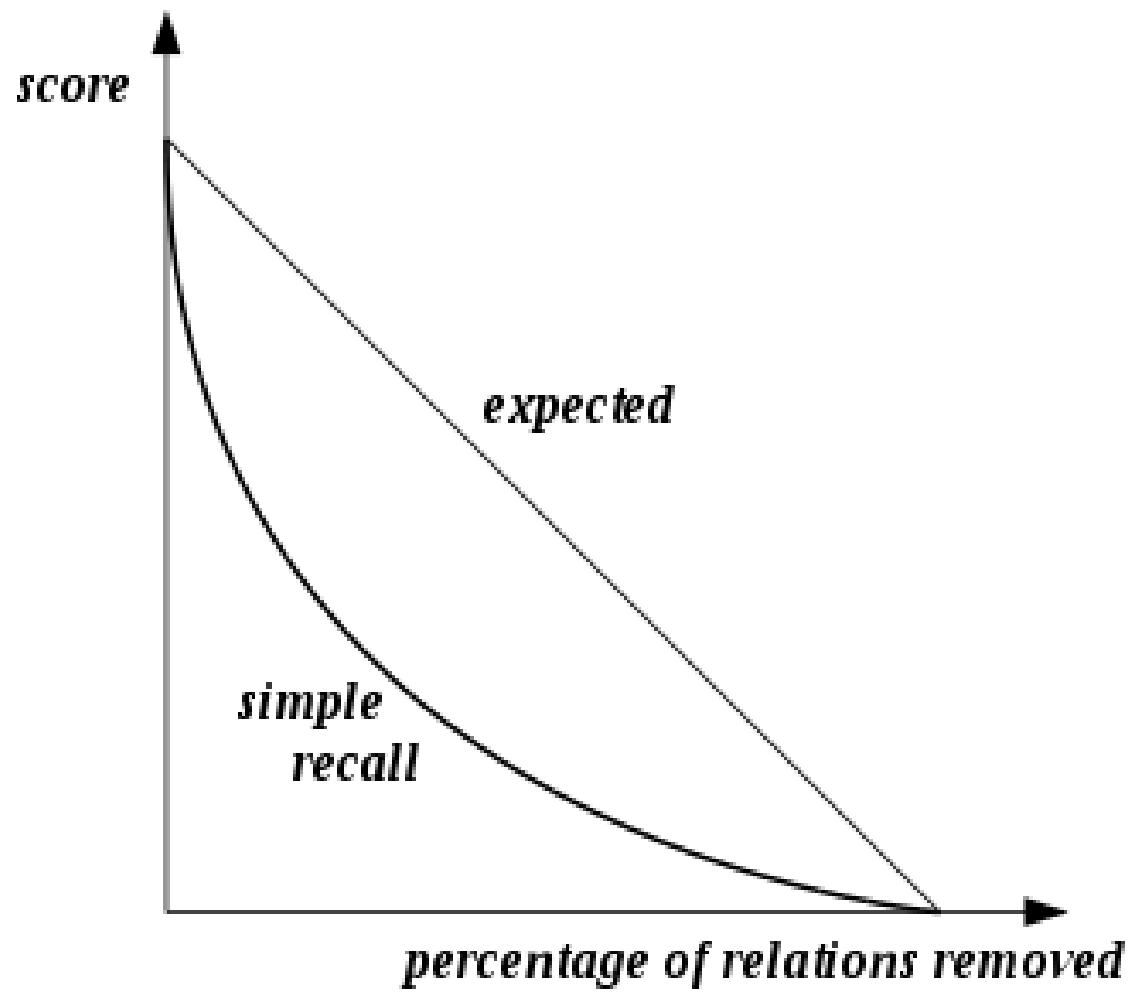

Figure 7: Recall behavior when removing information *vs.* ideal behavior.

### 2.6 Behavior of Existing Metrics

As stated in the work by Tannier and Muller (2008), a recall measure is expected to decrease in a linear way when the amount of information decreases. Otherwise, evaluation measures could have non-gradual changes that complicate comparisons of models. Besides, we expect the amount of information to grow roughly proportionally to the number of events in the text.

This behavior can be evaluated by comparing a given annotation with the same annotation where some of the temporal information is taken out.

Figure 7 shows how recall measures evolve when removing relations from a temporal graph, until no relations are present. It shows the different values of strict recall according to the proportion of relations kept in the graph, as well as the ideal $y = x$ line. This can be considered as an illustration of the consequences of an annotator "forgetting" to annotate some relations, with respect to an ideal reference.

This slope reveals two major drawbacks, both leading to a lack of stability of the metric:

- A non-linear progression of the curve does intuitively not correspond to what we expect from a good metric: for example, if a system A provides 60% more correct information than a system B, this system should get a 60% better recall.
- As will be shown later, the parabolic shape is due to the irregular redundancy and sparseness of human annotation. Then, two systems providing the same amount of correct information can get different recall values, depending on whether the human annotation of this information was redundant or not.

The Figure presents an idealized case; more thorough experiments were done on the whole TimeBank corpus (Pustejovsky et al., 2003) and explained in details in Section 6. For now, it is enough to note that the measure decreases in a parabolic way, since annotators roughly tag $O(n)$ relations where $n$ is the number of events, while inferred relations are in $O(n^2)$, the number of edges from $n$ nodes.

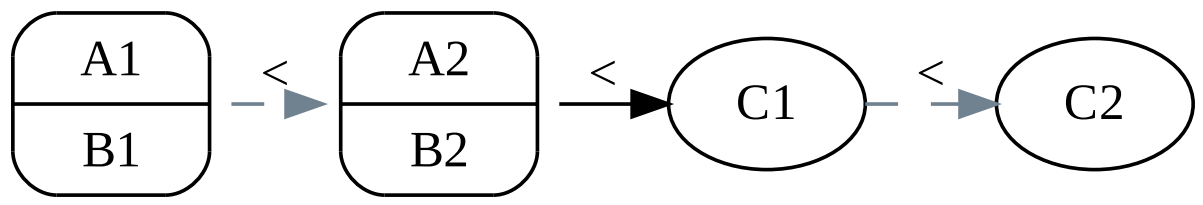

Figure 8: Endpoint graph with merges (same temporal information as Figure 4). Gray dashed arcs are trivial relations coming from the definitions of endpoints (C1 starts C, C2 ends C).

Full graphs contain redundant information, and recall thus decreases in artificial, irregular ways when some of it is removed. Our hypothesis is that working on minimal graphs will suppress redundancy and lead to a more controlled behavior.

Moreover, the fact that reference graphs contain $O(n^2)$ relations after closure biases the evaluation towards the larger texts (or at least texts containing large clusters of related events).

## 3. Proposed New Metric

When confronting a graph to a gold standard, a similarity measure is necessary. Many similarity measures exist between two graphs once a many-to-many correspondence is found between the nodes of both graphs (Sorlin, 2006).

Node matching, which is a major problem in graph comparison in general, is not difficult in our case, since we consider that both graphs annotate the same events or expressions[5].

A traditional similarity function between two graphs is the following (Sorlin, 2006):

$$sim(K, G) = \frac{f(K \sqcap_m G) - g(splits(m))}{f(K \cup G)}$$

where $K \sqcap_m G$ is the set of relations shared by both graphs according to the node matching function $m$, $K \cup G$ is the union between $K$ and $G$ relations, and $splits(m)$ the number of node splits imposed by the matching to obtain a graph mapped to the other (see examples later). Functions $f$ and $g$ depend on the types of graphs and applications.

But this kind of metrics is not appropriate for temporal relations, because the transitivity of relations implies different features; also, these metrics are symmetrical, whereas two distinct recall- and precision-like values are more desirable. We adapted the general idea of two functions for split nodes and relation similarity, and arrived at the algorithm described below.

### 3.1 Transitive Reduction of Endpoint Graph

To address the problem of finding minimal graphs and to take into account the relative importance of relations, we take inspiration from the work of Dubois and Schwer (2000) in two main ideas. First, graphs are saturated (*i.e.* temporal closure is applied), and are converted into endpoint graphs. Second, two nodes linked by an equality relation are merged together (this will help guarantee the uniqueness of the minimal graph, see below), a useful procedure on point-based graph (van Beek, 1992). Figure 8 presents the graph of Figure 3 after this transformation. The resulting point graph is saturated, by definition of the composition of event relations in Allen's algebra.

Recall that the graph we consider are built from "convex" annotations, *i.e.* there cannot be a "$<$ or $>$" relation between two points. We can keep relations $<$ and $\leq$ without loss of information, since all $>$ and $\geq$ can be obtained by symmetry.

5. If this is not the case, creating fictitious unlinked nodes in one graph or both is enough.

With these specifications, the graph boils down to the directed graph of a transitive relation where an edge between two points $x$ and $y$ means $x \leq y$. A coherent graph will thus be acyclic, since we collapse equal points into single nodes. It is important to note that as a consequence, no edge in the transitive closure can be labelled with the equality relation only. Thus we can see our problem as searching for the transitive reduction of a graph labelled with the transitive relation $\leq$ (but for which we keep the additional information that some edges can be more precisely labelled as $<$ instead of the disjunctive $\leq$). This is important because the minimal graph is the transitive reduction of the graph, and the transitive reduction of a directed acyclic graph is unique (Aho, Garey, & Ullman, 1972; La Poutré & van Leeuwen, 1988). The transitive reduction of a graph $G$ is by definition the subgraph corresponding to the minimal set of edges (with respect to inclusion) that has the same transitive closure as $G$, i.e. the minimal graph $G'$ such that $G'$ is a subgraph of $G$ and $G'^* = G^*$ where $G^*$ is the transitive closure of $G$. It is simply determined by $G^*/(G^* \circ G^*)$. Algorithm 2 details a simple computation of the transitive reduction while Figure 9 shows an illustration of the procedure on a simple transitive graph. Figure 10 shows the process from the initial endpoint graph with both $<$ and $\leq$ labels, to the minimal graph, *via* transitive reduction of the unlabelled graph.

**Algorithm 2** Transitive reduction simple computation

```
procedure COMPOSE(G)                         ▷ find relations inferable from others
    newRels={}
    base_rels= { x for x in G.edges() if x.relation() in {<, ≤}}

    for all one in base_rels do
        related ={x for x in G.edges() if x.source()=one.target() and x.relation() in {<, ≤} }

        for all other in related do
            relation = compose(one.relation(),other.relation())
            newRels.add(Edge(one.source(),other.target(),relation))
        end for
    end for
    return newRels
end procedure

procedure TRANSITIVE-REDUCTION(G)
    G = closure(G)
    non_min=compose(G)
    for all one in non_min do
        G.edges().remove(one)                ▷ remove relations deduced by composition
    end for
    for all one in G.edges() do

        if one!=before and one!=before_or_equals then
            G.edges().remove(one)    ▷ keep only <, ≤ and remove their symmetric relations
        end if
    end for
end procedure
```

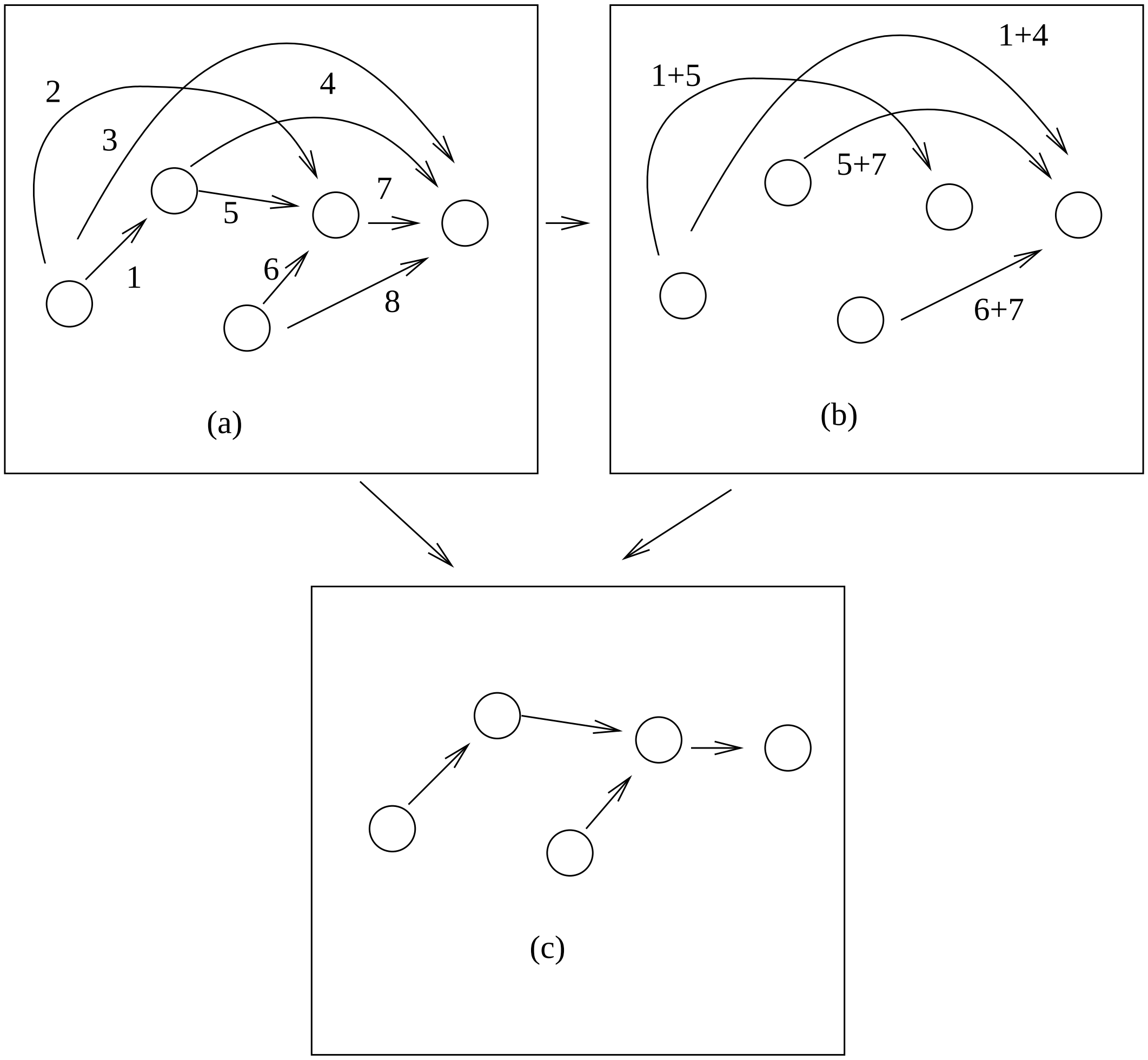

Figure 9: Transitive reduction of an acyclic graph; (a) is the initial closure of a transitive relation graph; (b) is the set of edges that can be obtained by composition of edges in (a), with examples of composition for each edge; (c) is the transitive reduction, the difference between (a) and (b).

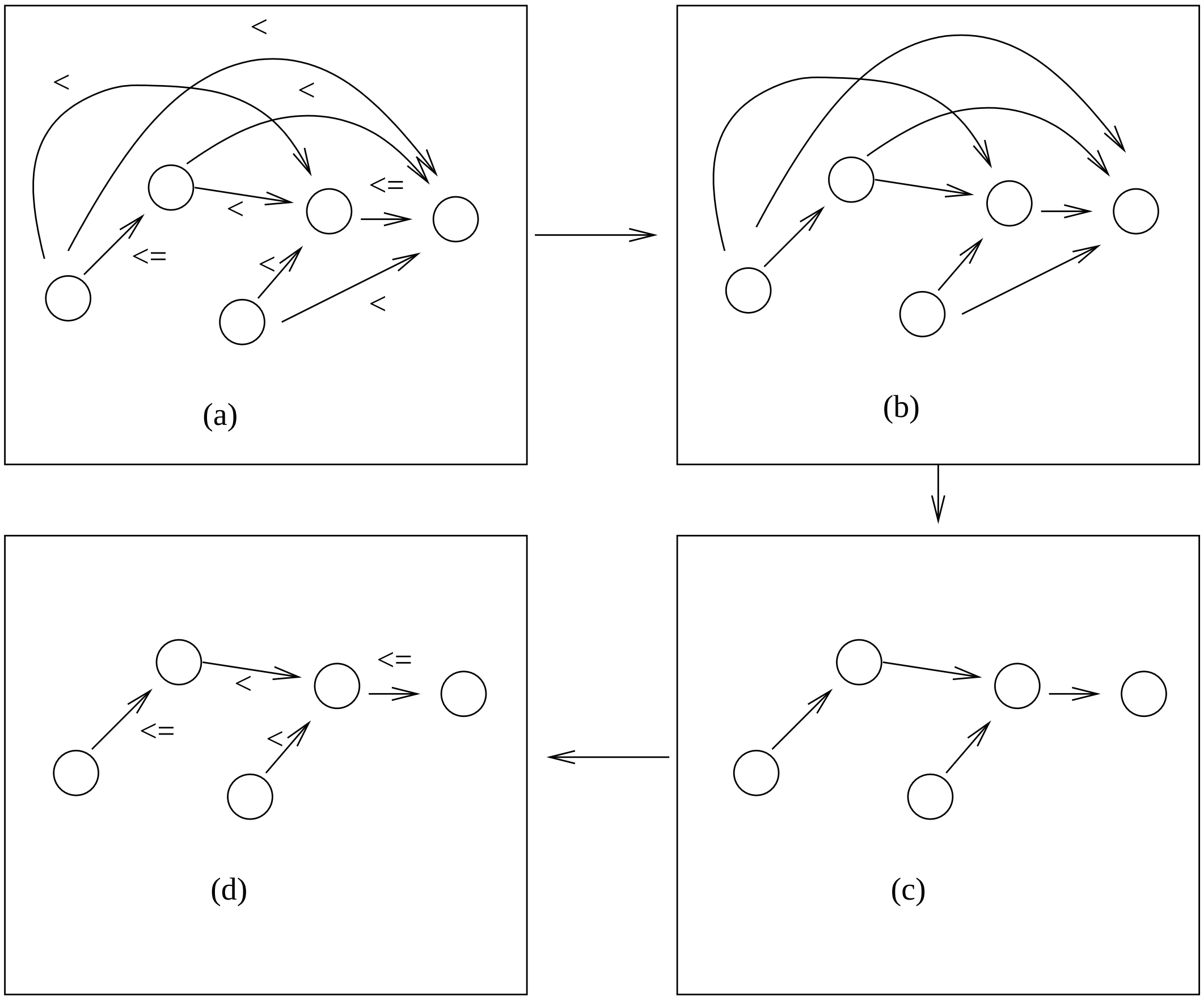

Figure 10: Transitive reduction of a point-based graph; (a) is the initial annotation transformed into a graph on events endpoints; (b) is the corresponding graph as if every label was $\leq$, (c) its transitive reduction and (d) the final minimal graph where the more precise initial information is reported.

We call:

- *Major* relations, the relations of the transitive reduction, or $G_{maj}$.
- *Minor* relations, the relations of temporal closure which are not present in the transitive reduction, *i.e.* $G^* - G_{maj}$.

Formally:

Let $G = \{(x, y, R)/R \in \{<, \leq\}\}$, the temporal point graph, saturated with respect to the relation $<$ and $\leq$. So $G^* = G$

Let $E(G) = \{(x, y)/\exists R, (x, y, R) \in G\}$, the unlabelled corresponding graph. The function $f$ which associates $(x, y, R)$ in $G$ to $(x, y)$ in $E(G)$ is an obvious bijection, as the original graph $G$ has at most one relation holding between any two vertices. Here $E(G) = f(G)$. Since $G$ is closed, so is $E(G)$.

Let $Proj(G', G) = \{(x, y, R) \in G/(x, y) \in G'\}$ the "projection" of an unlabelled graph into a labelled one. The function associating an edge $(x, y)$ in $G'$ to $(x, y, R)$ in $G$ is the inverse of $f$, $f^{-1}$. $Proj(G', G) = f^{-1}(G')$, and obviously $Proj(E(G), G)) = G$.

It is enough to prove that $E(G)$ (or $G$) is the graph of a transitive, acyclic relation to prove that $E(G)$ has a unique transitive reduction (Aho et al., 1972).

First, $E(G)$ is transitive: let $(x, y) \in E(G)$ and $(y, z) \in E(G)$ then we have $(x, y, <)$ or $(x, y, \leq) \in G$ and $(y, z, <)$ or $(y, z, \leq)$ in $G$ in any of the four possibilities we can infer either $(x, z, <)$ or $(x, z, \leq)$ is in $G$ (because $<$ is transitive, $\leq$ is transitive and $x < y \leq z$ or $x \leq y < z$ both imply $x < z$); in other words any composition of $<$ and $\leq$ is $<$, i.e. $((< \circ \leq) = (\leq \circ <) = (<))$ so $(x, z)$ is also in $E(G)$ and $E(G)$ is the graph of a transitive relation. This is the reason why we said that keeping a record of all $<$ relations while considering $G$ as a graph of $\leq$ does not change the graph property.

Second, $G$ is acyclic since $<$ is irreflexive, and $x < y \leq z$ implies $x < z$ so that the only way to have cycles in $G$ is if there are paths such as $x \leq y \leq z \leq ... \leq x$. But in that case we can infer that $x = y = z = ...$ and the nodes would have been merged beforehand. So $E(G)$ is also acyclic (it has exactly the same edges as $G$).

So, $E(G)$ is acyclic and transitive and thus admits a unique transitive reduction $E(G)_{tr}$.

Since the graphs $G$ and $E(G)$ have exactly the same edges, they necessarily have the same reductions, and thus both have a unique transitive reduction. We can project back the original relations of $G$ on $E(G)_{tr}$, $(Proj(E(G)_{tr}, G))$, to have a properly labelled reduction for $G$.

### 3.2 Temporal Recall and Precision

The idea is not to compare only minimal graphs. Temporal closure should be used as well. As we showed in Section 2.4, reference minor relations should still be rewarded if they are not redundant in the evaluated graph. However, they must carry a lower weight.

Minor relations are only to be considered in temporal recall, and not in precision. The reason is that recall evaluates the proportion of reference relations found by the system, and this system can find some minor relations without the major relations that produced them in the reference (see $B\, b\, D$ in $S_4$ example, Figure 6). On the opposite, precision evaluates the proportion of system relations that are in the reference graph, and minor (i.e. deducible) relations found by the system are by definition redundant.

Our recall-like measure is then a combination of two values. Given $K$ the reference graph and $G$ the evaluated graph:

- The major temporal recall is the rate of reference major relations ($K_{maj}$) found in $G^*$.
- The minor temporal recall is the rate of reference minor relations ($K^* - K_{maj}$) found in $G_{maj}$.

In the first case, the temporal closure is applied to $G$, since there is no reason to restrain the search of good relations in the evaluated graph. In the second case, only the transitive reduction $G_{maj}$ is considered; reference minor relations must be rewarded only if they are not minor in $G$ (case of $B\,b\,D$ in example $S_2$, Figure 6). $G$ minor relations have already been assessed through their major relations (case of $A\,b\,C$ in example $S_4$).

The final value of temporal recall is a weighted sum of the two figures.

The precision-like measure is a single value corresponding to the ratio between correct relations in $G_{maj}$ and its total number of relations. $G$ minor relations must not be considered at all for precision. Unlike for recall where reference major relations may not be retrieved by the system, precision necessarily considers major relations from the system. Minor relations are then all redundant.

In precision as well as in recall, a merge must be considered as a relation in some way, because it corresponds to an '=' relation.

### 3.3 Notations

We note:

- $G$ and $K$ the event graphs, $K$ being the reference (key);
- $G^{pt}$ and $K^{pt}$, the corresponding endpoint graphs, where equals points are merged.
- $(N, R)$ the set of nodes and relations of the graph $G$, and *mutatis mutandis* for the others.
- in the examples, only **non-trivial** relations are listed. **Trivial** relations are those involving two points of a same interval. For example, $A1 < A2$ in Figure 8 is trivial, and thus not considered. In all figures, trivial relations are dashed gray lines. Non-trivial relations and the node list are enough to find the full graph.
- The **temporal closure** of $G$ is noted $G^*$.
- The **transitive reduction** of $G$ is now noted $G_{maj}$.

### 3.4 Temporal Recall and Precision

Our new metrics of temporal recall and precision rely on the notions defined above. With respect to the recall value, we have shown it was important to distinguish "major" relations, *i.e.* relations that belong to the minimal graph, from the other, "minor" relations. That is why we suggest to compute recall in two steps. Precision is not concerned by this issue.

#### 3.4.1 Graph Values

Let's consider now $G^{pt}$ a set of merged nodes and relations $(N^{pt}, R^{pt})$ evaluated with respect to the reference $K^{pt} = (N^{pt}_K, R^{pt}_K)$. We need to take into account the number of relations expressed in the original point graph before nodes are merged (called value($G^{pt}$)), that is the number of equalities expressed in the merged nodes, $node_values(G^{pt})$, plus the number of relations $relation_value(G^{pt})$. Then:

$$\begin{array}{rclcl} value(G^{pt}) & = & node_value(G^{pt}) & + & relation_value(G^{pt}) \\ & = & \sum_{n_i \in N^{pt}} (|m(n_i)| - 1) & + & |R^{pt}| \\ & = & |N| \times 2 - |N^{pt}| & + & |R^{pt}| \end{array}$$

Here, $m(n_i)$ the number of points merged into a node ($m$ points into a node correspond to $m-1$ merges). Similarly, we can compute the value of the reference graph.

The pairing of nodes in the evaluated graph and the reference graph must be then taken into account. Let's call a **"split"** the operation of mapping one node of the reference graph to many nodes in the evaluated one, and a **"conflation"**, the converse operation. To map both graphs, we first need to split any node of the reference that is not in the evaluated graph into nodes that are in it and maybe some extra nodes, and then conflate, if necessary, the remaining nodes to evaluated nodes (see Sections 4 and 5 for illustration). A split is like a '=' relation provided by the reference but not by the evaluated graph. A conflation is a '=' wrongly predicted by the evaluated graph.

Therefore, the number of correct answers in an evaluated graph, or *correct value* $v_c$, is the number of correct '=' and '<' relations[6]. The number of correct '=' is "node value - nb of conflations", the number of correct '<' is "relation value - wrongly predicted relations (errors)".

A simple way of calculating the number of splits necessary to match two graphs G1 and G2 is to count for each node x of G1 the number $i$ of different nodes of G2 that intersect with x. If $i$ is one, the node is mapped directly, otherwise $i-1$ splits are needed. Thus:

$$|split(K^{pt}, G^{pt})| \quad = \sum_{x \in N_K^{pt}} |\{y \in N^{pt} / x \cap y \neq \emptyset\}|$$

The number of conflation necessary to match G1 to G2 is also the number of splits necessary to match G2 to G1, so:

$$|conflation(K^{pt}, G^{pt})| \quad = |split(G^{pt}, K^{pt})|$$

$$\begin{aligned} v_c(G^{pt}) &= \text{correct equalities} + \text{correct precedences} \\ &= (\text{node_value}(G^{pt}) - \text{conflations})) + (\text{relation_value}(G^{pt}) - \text{errors}) \\ &= v(G^{pt}) - (|conflation(K^{pt}, G^{pt})| + \text{errors}) \\ &= v(G^{pt}) - (|split(G^{pt}, K^{pt})| + |R - R_K|/|R|) \end{aligned}$$

### 3.4.2 Temporal Recall and Precision

A precision value deals with errors (incorrect relations) and conflations through the correct value $v_c$. On the other hand, a recall value must take into account misses (reference relations missed by the graph) and splits. For simplicity, let's note now $K = K_K^{pt}$ the point-based reference, and $G = G^{pt}$ the graph to evaluate. $K_{maj}$ and $G_{maj}$ are the transitive reductions of these. We define:

- *Major temporal recall* $R_t(G)$ is the number of reference major relations found by $G^*$.

$$\begin{aligned} R_t(G) &= \frac{v(K_{maj}) - (splits + misses)}{v(K_{maj})} \\ R_t(G) &= \frac{v(K_{maj}) - (|split(K, G)| + |(R_K - R^*)/R_K|)}{v(K_{maj})} \end{aligned}$$

  where $misses$ is the number of relations in $K_{maj}$ missed by $G^*$ and $splits$ the number of splits (a split is a missed '=' relation).

6. We consider of course only non trivial relations.

- Symmetrically, *Temporal precision* $TP(G)$ is the ratio between the *correct value* of $G_{maj}$, $v_c(G_{maj})$, and its full *value* $v(G_{maj})$.

$$TP(G) = \frac{v(G_{maj}) - (conflations + errors)}{v(G_{maj})}$$
$$TP(G) = \frac{v_c(G_{maj})}{v(G_{maj})}$$

But, as we already stated, minor relations should also be taken into account. If we consider only major recall, systems that find only minor (but correct) relations are disadvantaged. This is the case of system $S_2$ in Figure 6 ($B\ b\ D$ is correct and minor). That is why we add a *minor temporal recall*:

- *Minor temporal recall* $r_t(G)$ is the proportion of reference minor relations ($K^* - K_{maj}$) found by $G_{maj}$. Minor relations are seeked in the minimal evaluated graph. Indeed, as we already said, comparing two relations from non-minimal graphs is redundant, since major relations that "produced" them have already been taken into account.

- *Full temporal recall* $TR$ could be defined as a value pair $(R_t(G), r_t(G))$, or preferably as a combination:

$$TR(G) = R_t(G) + \frac{1}{v(K_{maj})} r_t(G) \tag{2}$$

  With this formula, we ensure that one single major relation is better than all minor ones and that the recall can not exceed 1 (see next Section).

The symmetrical precision property, *i.e.* the proportion of system minor relations existing in the reference, is always null. System minor relations are, by definition, always redundant. That is why temporal precision does not need a two-fold figure.

A tool for computing different precision and recall values (strict, relaxed, core, as well as this new one) is made available with this paper[7]. It accepts several input formats, including TimeML. All computed data are also available, with instructions to reproduce the experiments presented at the end of this paper.

#### 3.4.3 SYNTHESIS

Evaluating an interval graph $G$ against a gold reference $K$ is achieved through the following steps:

1. Perform transitive closure on both $G$ and $K$.

2. Convert both graphs into endpoint graphs.

3. Merge equality relations into single nodes.

4. Perform transitive reduction.

5. Compute values of temporal recall and precision.

7. http://www.irit.fr/~Philippe.Muller/resources.html

The most costly operation here is the first one, and is common to the standard evaluation procedure; its time is in $O(|N|^3)$ in the worst case with the algorithm 1 page 379, N being the set of events of the largest graph between G and K. The other are straightforwardly linear in the number of relations of the graph (conversion to endpoints, merge equalities in the closed graph), or at worse in $O(|N|^2)$ for both the transitive reduction (one composition of relations in the graph and set difference), and the computation of the number of splits and conflations. Overall, the whole procedure is dominated by the first step, which is common to all evaluation procedures for this task. Our proposition thus does not change the worst-case complexity of the evaluation procedure.

#### 3.4.4 BOUNDARIES

**Temporal precision.** Temporal precision is a value between 0 and 1 since $v_r(G) \leq v(G)$ (errors and conflations are zero or positive).

**Temporal recall.** Major recall is between 0 and 1.

- If 1, then minor recall $r_t(G) = 0$, because all relations in $G_{maj}$ are already in $K_{maj}$ (and then cannot be in $K^* - K_{maj}$). In this case temporal recall cannot exceed 1.
- If not 1,
  $R_t(G) \leq (v(K_{maj}) - 1)/v(K_{maj})$. Yet, $\frac{1}{v(K_{maj})} r_t(G) < \frac{1}{v(K_{maj})}$, and the full temporal recall stays below 1.

Thus $0 \leq TR(G) \leq 1$.

In order to make the measure clearer, we will present the metric in detail by developing a basic example with a transitive closure made only of simple relations (the 13 basic Allen relations), then we will turn to the more general case of a closure made only of *convex* temporal relations (disjunctions of neighboring Allen relations).

#### 3.4.5 SIMPLE EXAMPLES

Temporal recall and precision as described above lead to the expected values for the sample graphs pictured in Figure 6 and analyzed in Section 2.4.

Figure 11 recalls these graphs and details the temporal recall values for each of them (given that $v(K_{maj}) = 3$). As for precision, for each graph $S_i$, $TP(S_i) = 1$.

An attentive reader would note that a system providing *only* minor relations, $A < C$, $A < D$ and $B < C$, *i.e.* half of all deducible relations, would get a recall of only 0.33, which could seem unfair. However, even if three relations are present, three others are still missing (the same number of missing relations as if the graph had no relation at all). Moreover, the fact that all minor relations does not get a better score than one major relation is the condition for the boundaries not to go over 1. Finally, this case is rare and does not affect the general behavior of the metric.

More sophisticated examples are provided in the following sections.

## 4. First Simple Case: Non-Disjunctive Allen Relations

Consider the sample graph $K1$ made of Allen non-disjunctive relations (see also the graph in Figure 12):

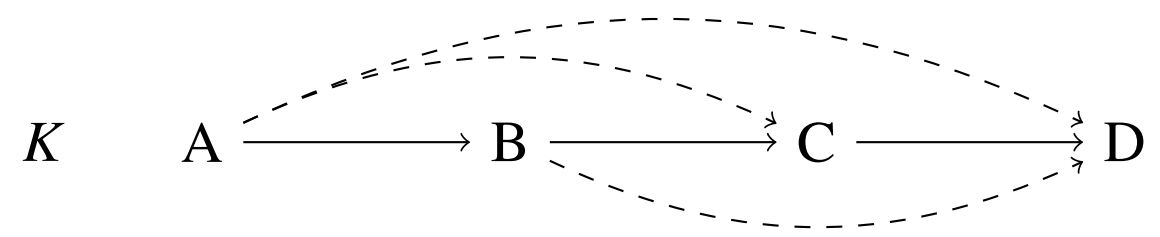

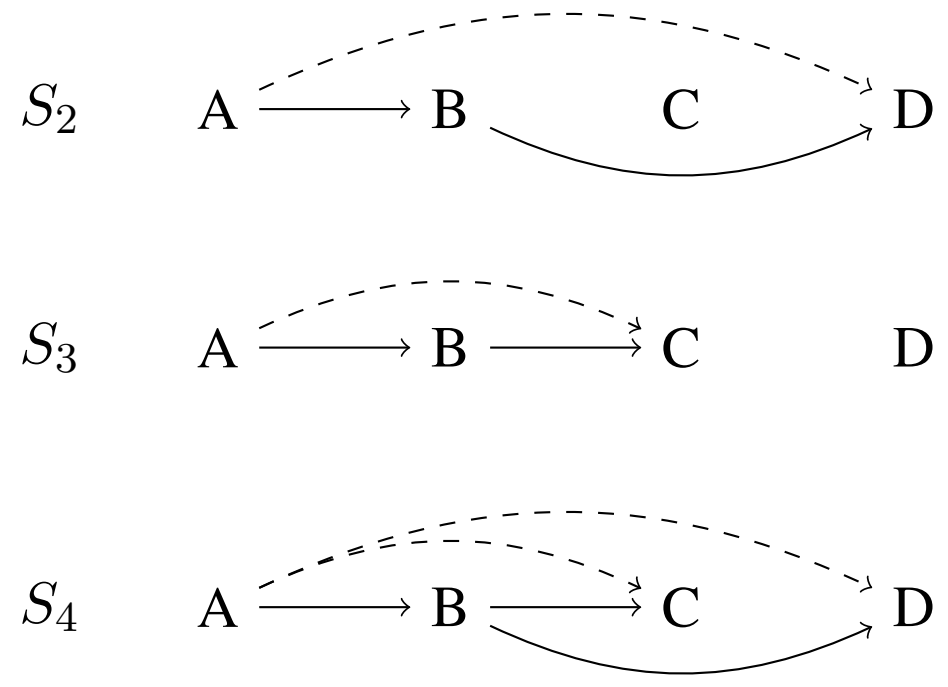

$TR(S_1) = \frac{3-(1+0)}{3} + \frac{0}{3} = 0.67$
(2 major relations, 1 missing)

$TR(S_2) = \frac{3-(2+0)}{3} + \frac{1}{3\times 3} = 0.44$
(1 major and 1 minor relations)

$TR(S_3) = \frac{3-(1+0)}{3} + \frac{0}{3} = 0.67$
(2 major and 1 redundant minor relations)

$TR(S_4) = \frac{3-(1+0)}{3} + \frac{1}{3\times 3} = 0.77$
(2 major and 1 non-redundant minor relations)

Figure 11: Temporal recall of simple graphs, compared to reference $K$.

$K1 =$

| | $A$ | $B$ | $C$ | $D$ | $E$ | $F$ |
|---|---|---|---|---|---|---|
| $A$ | | $s$ | $bi$ | $b$ | $s$ | $b$ |
| $B$ | | | $bi$ | $m$ | $s$ | $m$ |
| $C$ | | | | $b$ | $b$ | $b$ |
| $D$ | | | | | $f$ | $e$ |
| $E$ | | | | | | $fi$ |

The conversion into relations between endpoints leads to the following graph, where an event A is split into A1 and A2. Edges are not labelled since only relation '<' is considered at this point.

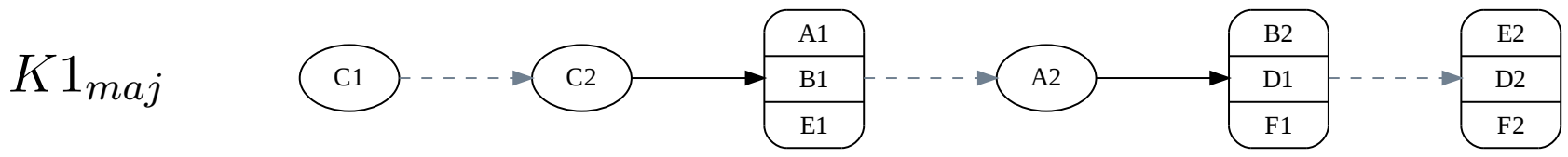

Figure 12: Reference with simple relations, $K1_{maj}$.

Note that merging equal nodes is not equivalent to labelling arcs with '=', because in the latter case the minimal graph would not be unique any more. For example, the necessary choice between $A1 < A2$, $B1 < A2$ and $E1 < A2$ would lead to three equivalent but different graphs.

Consider now that the reference $K1$ is compared to the following graph $G1$ (Figure 13; bold edges are correct relations, thin edges are wrong relations, dashed edges are trivial relations, *i.e.* involving two endpoints of the same interval).

Following the notations defined in Section 3.3:

- The list of nodes for graphs $K1_{maj}$ and $G1_{maj}$ are:

$G1_{maj}$

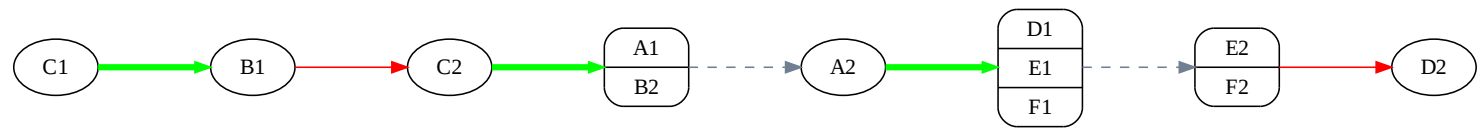

Figure 13: Evaluated graph with simple relations, $G1_{maj}$.

  - for $K1_{maj}$ : $N1 = \{C1, C2, (A1, B1, E1), A2, (B2, D1, F1), (E2, D2, F2)\}$ : 6 nodes
  - for $G1_{maj}$ : $N2 = \{C1, B1, C2, (A1, B2), A2, (D1, E1, F1), (E2, F2), D2\}$ : 8 nodes

- Non-trivial relations are:
  - for $K1_{maj}$ : $R1 = \{[C2; (A1, B1, E1)], [A2; (B2, D1, F1)]\}$ : 2 relations
  - for $G1_{maj}$ : $R2 = \{[C1; B1], [B1; C2], [C2; (A1, B2)], [A2; (D1, E1, F1)], [(E2, F2); D2]\}]$ : 5 relations

- The temporal closures $Gi^*$ are computed and listed below (bold relations are those added from the minimal graph, *i.e.* $K1^* - K1_{maj}$ and $G1^* - G1_{maj}$). Relations on the same line share their arguments (at least one point in common for each argument). Figures 14 and 15 also represent the closures.

| $K1^*$ | $G1^*$ |
|---|---|
| $[C2; (A1, B1, E1)]$ | $[C2; (A1, B2)]$ |
| $[A2; (B2, D1, F1)]$ | $[A2; (D1, E1, F1)]$ |
| $[\mathbf{C1}; (\mathbf{A1}, \mathbf{B1}, \mathbf{E1})]$ | $[C1; B1]$ |
| $[\mathbf{C1}; \mathbf{A2}]$ | $[\mathbf{C1}; \mathbf{A2}]$ |
| $[\mathbf{C1}; (\mathbf{B2}, \mathbf{D1}, \mathbf{F1})]$ | $[\mathbf{C1}; (\mathbf{A1}, \mathbf{B2})]$ |
| | $[\mathbf{C1}; (\mathbf{D1}, \mathbf{E1}, \mathbf{F1})]$ |
| $[\mathbf{C1}; (\mathbf{E2}, \mathbf{D2}, \mathbf{F2})]$ | $[\mathbf{C1}; \mathbf{D2}]$ |
| | $[\mathbf{C1}; (\mathbf{E2}, \mathbf{F2})]$ |
| $[\mathbf{C2}; \mathbf{A2}]$ | $[\mathbf{C2}; \mathbf{A2}]$ |
| $[\mathbf{C2}; (\mathbf{B2}, \mathbf{D1}, \mathbf{F1})]$ | $[\mathbf{C2}; (\mathbf{D1}, \mathbf{E1}, \mathbf{F1})]$ |
| $[\mathbf{A2}; (\mathbf{E2}, \mathbf{D2}, \mathbf{F2})]$ | $[\mathbf{A2}; (\mathbf{E2}, \mathbf{F2})]$ |
| | $[\mathbf{A2}; \mathbf{D2}]$ |
| $[\mathbf{C2}; (\mathbf{E2}, \mathbf{D2}, \mathbf{F2})]$ | $[\mathbf{C2}; (\mathbf{E2}, \mathbf{F2})]$ |
| | $[\mathbf{C2}; \mathbf{D2}]$ |
| | $[B1; C2]$ |
| | $[(E2, F2); D2]$ |
| | $[\mathbf{B1}; \mathbf{A2}]$ |
| | $[\mathbf{B1}; (\mathbf{D1}, \mathbf{E1}, \mathbf{F1})]$ |
| | $[\mathbf{B1}; (\mathbf{E2}, \mathbf{F2})]$ |
| | $[\mathbf{B1}; \mathbf{D2}]$ |
| | $[(\mathbf{A1}, \mathbf{B2}); (\mathbf{D1}, \mathbf{E1}, \mathbf{F1})]$ |
| | $[(\mathbf{A1}, \mathbf{B2}); (\mathbf{E2}, \mathbf{F2})]$ |
| | $[(\mathbf{A1}, \mathbf{B2}); \mathbf{D2}]$ |

- The pairing of nodes and the list of splits and conflations needed to match both sets of nodes is also computed (see also Figure 16):

$K1^*$

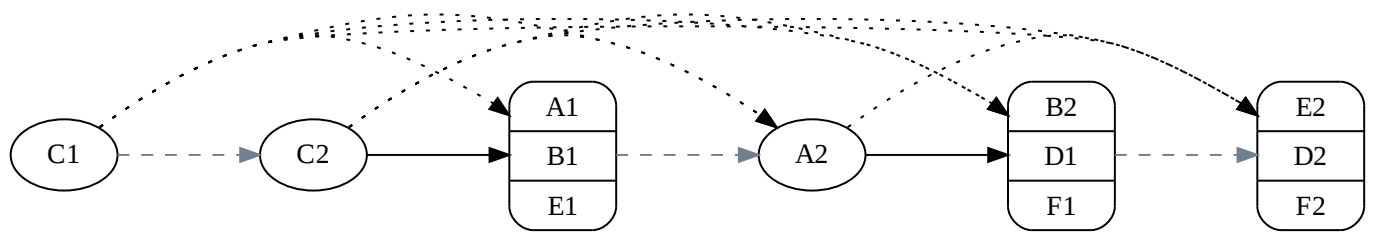

Figure 14: Temporal closure $K1^*$. Dotted relations represent $K1^* - K1_{maj}$.

$G1^*$

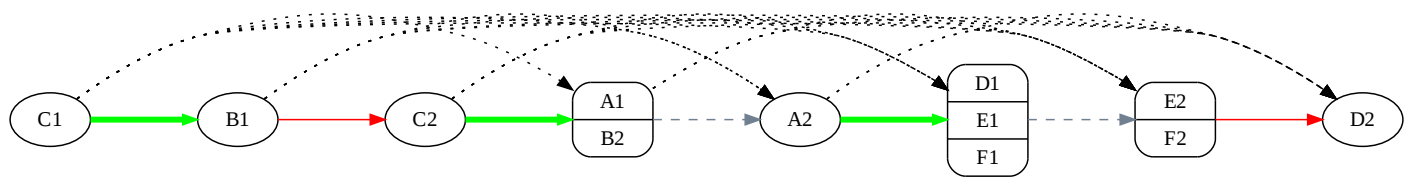

Figure 15: Temporal closure $G1^*$. Dotted relations represent $G1^* - G_{maj}$.

- $(A1, B1, E1)_{K1}$: 2 splits ("breaking" of 2 equality relations); $(B2, D1, F1)_{K1}$: 1 split ($D1$ and $F1$ stay together in $G1$); $(E2, D2, F2)_{K1}$: 1 split
- $(A1, B2)_{G1}$: 1 conflation (joining two nodes); $(D1, E1, F1)_{G1}$: 1 conflation (joining $E1$ to the two others); $(E2, F2)_{G1}$: nothing (already together in $K1$)
- Total: 4 splits and 2 conflations.

As stated above, an edge is *correct* if the relation is correct for at least one pair of points from both nodes. For example, relation $\{A2\}_{G1} \rightarrow \{D1, E1, F1\}_{G1}$ is correct because $\{A2\}_{K1} \rightarrow \{B2, D1, F1\}_{K1}$, even if the sub-relation $A2 < E1$ is not true. This latter relation will be penalized anyway because a split is necessary for matching the graphs.

These definitions lead to the following values;

- Graph values:
  - $v(K1_{maj}) = node\ value + relation\ value = 6 + 2 = 8$
  - $v(G1_{maj}) = 4 + 5 = 9$
- Correct value of G1:

$$\begin{aligned} v_c(G1_{maj}) &= (node\ value - conflations) + (relation\ value - errors) \\ &= v(G1_{maj}) - (conflations + errors) \\ &= 9 - (2 + 2) = 5 \end{aligned}$$

- Major temporal recall:

$$\begin{aligned} R_t(G1) &= \frac{v(K1_{maj}) - (misses + splits)}{v(K1_{maj})} \\ &= \frac{8 - (0 + 4)}{8} = 0.5 \end{aligned}$$

- Minor temporal recall: $r_t(G1) = \frac{2}{8} = 0.25$
  (this corresponds to $[C1; \{A1, B1, E1\}]$ and $[C2; (B2, D1, F1)]$).

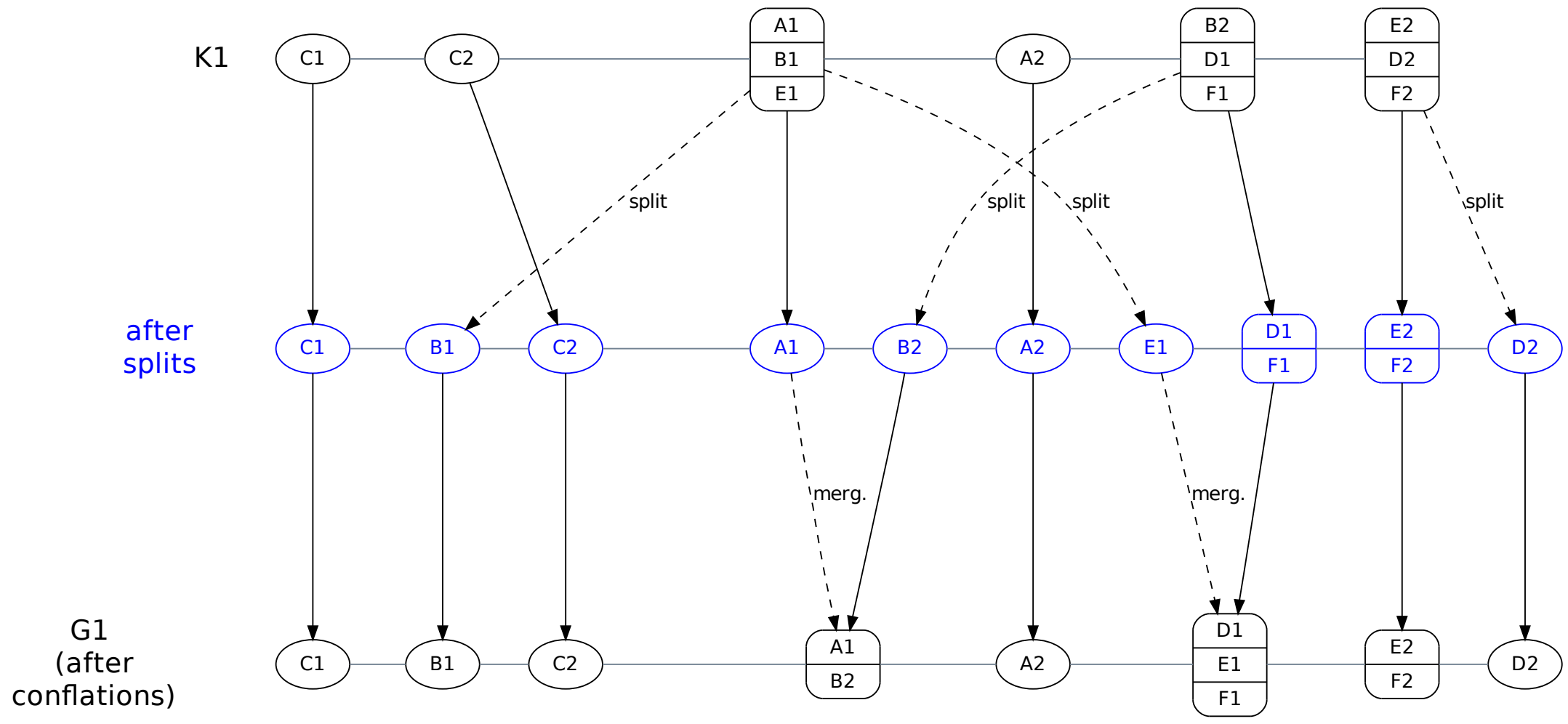

Figure 16: Split and conflation operations between $K1$ and $G1$.

- Full temporal recall:

$$\begin{aligned} TR(G1) &= (R(G1), r(G1)) \\ &= (0.5, 0.25) \\ \text{or } TR(G1) &= R_t(G1) + \frac{1}{v(K_{maj})} r_t(G1) \\ &= 0.5 + \frac{0.25}{8} = 0.53 \end{aligned}$$

- Temporal precision

$$\begin{aligned} TP(G1) &= \frac{v_c(G_{maj})}{v(G_{maj})} \\ &= \frac{5}{9} = 0.56 \end{aligned}$$

## 5. Disjunctive Convex Relations

We now apply the metric to sets of convex relations. Convex relations are a set of relations that are *conceptual neighbors*, that is they encode relations that may be vague but in which intervals endpoints can only be in convex subsets of the time-line (see Section 2.2).

Building the minimal graph follows the same procedure as explained above (transitive reduction). The measures do not differ at all if we choose a *strict* scoring scheme (see Section 2.3). But in a *relaxed* scheme, with disjunctions, it becomes necessary to apply a weighting procedure; a response is no longer assigned a binary value (0 or 1), but for example one of the values in Table 3, in a manner similar to what was done in TempEval (Verhagen et al., 2007).

As shown in that table, the relaxed measure has an effect on *misses* values and on $v_c$ (a relation can get a half-point), but also on conflation and split values (where half-points are also possible).

| *Rel.* | ↗ *Repres.* | = (A / B) | < (A →(<) B) | ≤ (A →(<, =) B) | > (B →(<) A) | ≥ (B →(<, =) A) |
|---|---|---|---|---|---|---|
| = | A / B | 1 | 0 | 0.5 *("$\frac{1}{2}$-split")* | 0 | 0.5 *("$\frac{1}{2}$-split")* |
| < | A →(<) B | 0 | 1 | 0.5 *(disjunction)* | 0 | 0 |
| ≤ | A →(<, =) B | 0.5 *("$\frac{1}{2}$-confl.")* | 0.5 *(disjunction)* | 1 | 0 | 0.5 *(disjunction)* |
| > | B →(<) A | 0 | 0 | 0 | 1 | 0.5 *(disjunction)* |
| ≥ | B →(<, =) A | 0.5 *("$\frac{1}{2}$-confl.")* | 0 | 0.5 *(disjunction)* | 0.5 *(disjunction)* | 1 |

Table 3: Example weights for relaxed measures.

Consider a new reference $K2$ with convex relations (see also Figure 17):

| | $A$ | $B$ | $C$ | $D$ | $E$ | $F$ |
|---|---|---|---|---|---|---|
| $A$ | | $s$ | $\{bi, mi, oi, fi, =, f, di, si\}$ | $b$ | $s$ | $b$ |
| $B$ | | | $\{di, si, oi, mi, bi\}$ | $m$ | $s$ | $m$ |
| $C$ | | | | $b$ | $\{d, s, o, m, b\}$ | $b$ |
| $D$ | | | | | $f$ | $\{s, =, si\}$ |
| $E$ | | | | | | $\{di, fi, o\}$ |

The characteristics of K2 are:

- 7 nodes for 6 events (5 equality relations)
- 2 relations: $[C2 \leq A2], [A2 < \{B2, D1, F1\}]$
- $K2^* =$ $[C2 \leq A2]$, $[A2 < \{B2, D1, F1\}]$, $\mathbf{C1 < A2}$, $\mathbf{C1 < \{B2, D1, F1\}}$, $\mathbf{C1 < \{E2, D2\}}$, $\mathbf{C1 < F2}$, $\mathbf{C2 < \{B2, D1, F1\}}$, $\mathbf{C2 < \{E2, D2\}}$, $\mathbf{C2 < F2}$, $\mathbf{\{A1, B1, E1\} < F2}$, $\mathbf{A2 < \{E2, D2\}}$, $\mathbf{A2 < F2}$
- $v(K2_{maj}) = (12 - 7) + 2 = 7$.

Let us consider an evaluated graph $G2$ (Figure 18) which has:

- 10 nodes for 6 events (2 equality relations)
- 5 relations: $[C1 \leq D1], [D2 < C2], [C2 \leq A2], [\{A1, B1\} \leq E1], [A2 < \{B2, F1\}]$
- $G2^* = [C1 \leq D1]$, $[D2 < C2]$, $[C2 \leq A2]$, $[\{A1, B1\} \leq E1]$, $[A2 < \{B2, F1\}]$, $[\mathbf{C1 < D2}]$, $[\mathbf{C1 < A2}]$, $[\mathbf{C1 < \{B2, F1\}}]$, $[\mathbf{C1 < F2}]$, $[\mathbf{D1 < C2}]$, $[\mathbf{D1 < A2}]$, $[\mathbf{D1 < \{B2, F1\}}]$, $[\mathbf{D1 < F2}]$, $[\mathbf{D2 < A2}]$, $[\mathbf{D2 < \{B2, F1\}}]$, $[\mathbf{D2 < F2}]$, $[\mathbf{C2 < \{B2, F1\}}]$, $[\mathbf{C2 < F2}]$, $[\mathbf{A2 < F2}]$, $[\mathbf{\{A1, B1\} < F2}]$, $[\mathbf{\{A1, B1\} < E2}]$
- $v(G2_{maj}) = (12 - 10) + 5 = 7$.

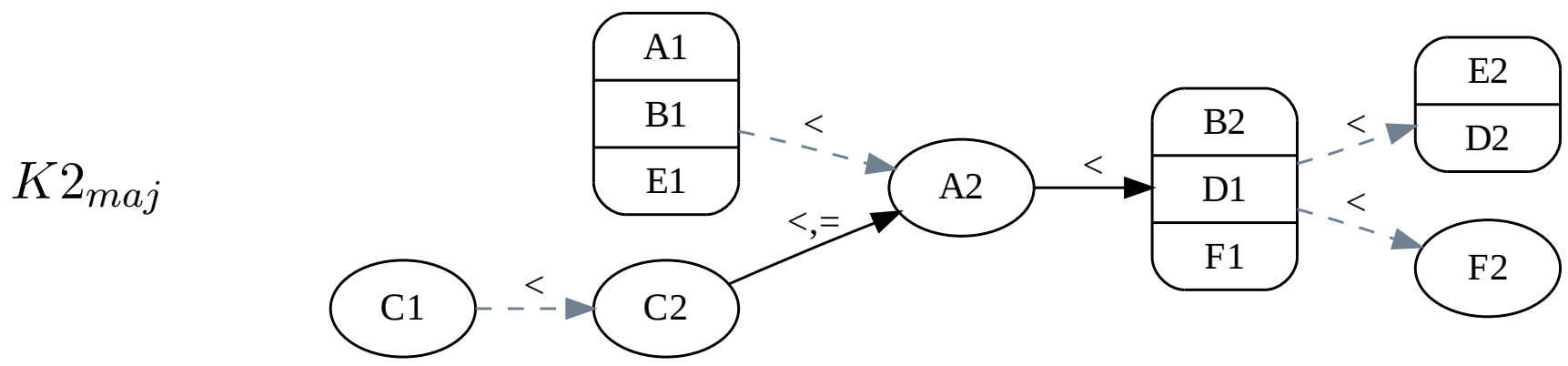

Figure 17: Reference with convex relations, $K2_{maj}$.

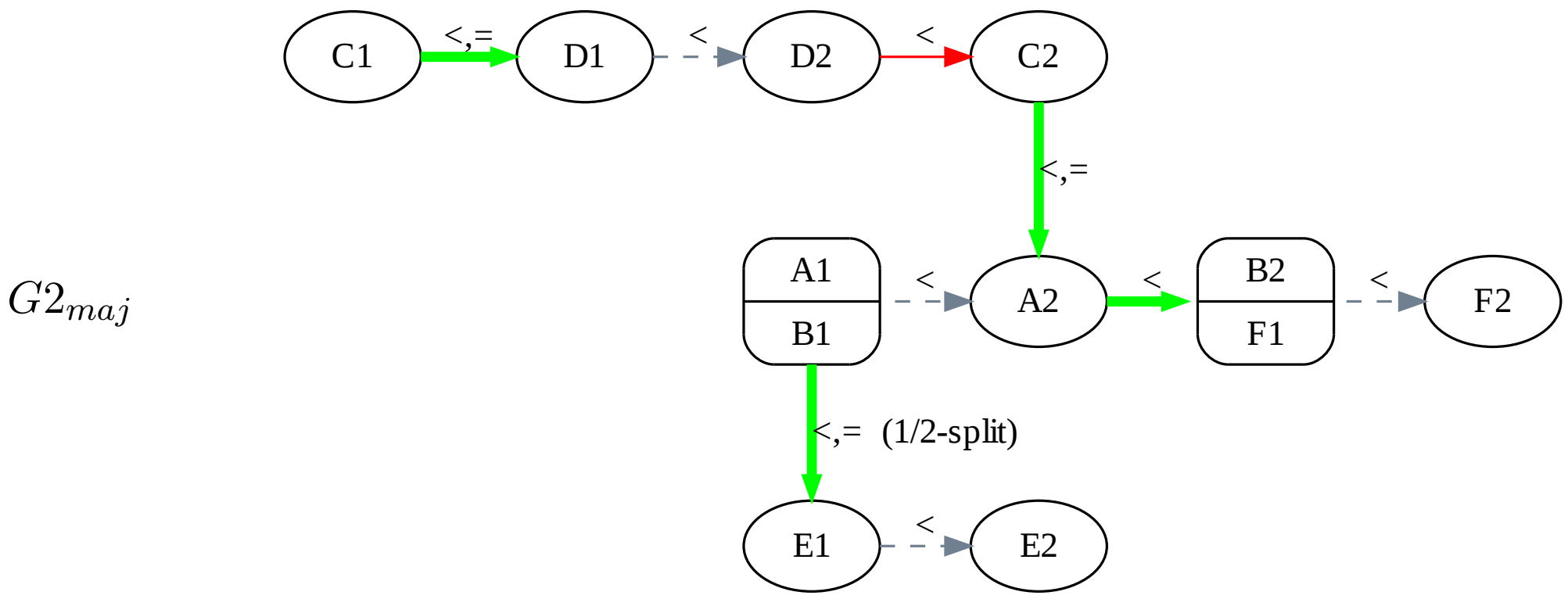

Figure 18: Evaluated graph with convex relations, $G2_{maj}$.

There is no conflation and the number of splits is 2.5. The "half-split" comes from the fact that E1 may be equal to $\{A1, B1\}$, because of the $\leq$ edge.

The application of the same measures leads to the following values:

$$R_t(G2) = \frac{7 - (0 + 2.5)}{7} = 0.64$$

$$r_t(G2) = 0$$

$$TR(G2) = 0.64$$

$$v_c(G2) = (12 + 2 * 1/2) + 2 = 15$$

The two half-points in $v_c(G2)$ hold for relations $C1 \leq D1$ (instead of $C1 < D1$ in the reference) and $B1 \leq E1$ (instead of $B1 = E1$ in the reference).

As an example, and following Table 3, we can see that $E1$ leaving the group $\{A1, B1, E1\}$ costs a half-point to the recall value, while relation $B1 \leq E1$, correct but imprecise, gets a half-point of precision. Both results seem logical behaviors of the measure.

## 6. Experiments

Our push for a change of evaluation measures of temporal annotation graphs was motivated in the beginning of the paper by a few examples where we show some undesirable side effects of common evaluation procedures. We will now present a more thorough methodology, aiming to evaluate the evaluation procedure itself. We used two kinds of data for that purpose. Obviously, we can test and compare measures on the freely available temporal annotated data provided in the TimeBank corpus. We also introduce a set of artificially built temporal graphs, on which we can control a few relevant parameters. TimeBank annotations are rather heterogenous, because human annotators make mistakes, forget relations, or introduce inconsistencies, thus creating a fair amount of noise. Besides, we wanted to test other aspects of temporal annotation that are easier to generate from scratch. One of this factor is the amount of information that is present in an annotation. We have seen that the result of an annotation can be more or less underspecified: a relation between events can be a simple, precise relation, or a disjunction of simple relations. The size of a graph is also probably important, but it is somewhat hidden in some human annotations, because the really relevant object when considering inferred information is a connected component. Small subgraphs with only a few nodes do not allow for a lot of inferences, and some of the bias we want to study probably happens only above a certain threshold. In human annotations, the size of subgraphs can vary a lot, and it is common for events to be related only to one or two other events, even after the graph has been enriched through inference procedures.

The main experiment we performed to study the behavior of commonly adopted measures and our own proposal is based on a comparison of an annotation to a "weakened" version of itself. For each graph, we remove a portion of event-event relations at random, and we use the measures to estimate the loss of information. Obviously, this is only interesting for recall-like measures, precision is not affected. We vary the amount of removed information in steps, until all relations are vague: eventually, the universal disjunction hold between any two events of the text. We designed another experiment to watch also the behavior of precision measures : instead of removing relations, we disturbed the graph by changing a simple relation to another one (simulating an "error" in the annotation, in a way). This is meaningful only if we keep the graph consistent at the same time.

In what follows, we present our results on the synthetic graphs (6.1) and then on the TimeBank data (6.2).

### 6.1 Artificial Graphs

In order to control the amount of information present in a temporal graph, we have built a set of artificial temporal graphs in the following way:

- For a given number of events $E$, a temporal graph is built by randomly choosing a set of $E$ integer pairs $(b_i, e_i)$, within a given range $N$, and with a level of indeterminacy $I$, that is a width around the boundaries $b_i$ and $e_i$.
- For each pair of events $(i, j)$, we determine a temporal relation between their intervals, considering that the endpoints of $i$ can lie anywhere in the interval $[b_i - I/2, b_i + I/2]$ and $[e_i - I/2, e_i + I/2]$, and the endpoints of $j$ can lie anywhere in the interval $[b_j - I/2, b_j + I/2]$ and $[e_j - I/2, e_j + I/2]$.
- The graph is closed, leading to $N = (E \times (E - 1))/2$ relations.

Events are considered as intervals with some uncertainty about their endpoints, to be comparable to a graph built from a text, and to give rise to possibly disjunctive relations between the generated events. This graph can thus contain any kind of convex (possibly disjunctive) Allen relations. An example of such generated events is provided by Figure 19. By varying $N$ and $I$, we can control the amount of vagueness of the information put in the graph. The smaller $N$ is, the tighter the model we generate, and the most likely it is that vague boundaries intersect, creating disjunctive relations. A larger $I$ also contributes to a vaguer representation. We quantify the amount of vagueness $v$ by how much the relations are disjunctive in the graph:

$$v = \frac{\sum_{r \in edges(G)} 1 - \frac{1}{|r|}}{|edges(G)|}$$

where $r$ is an edge in the graph and $|r|$ is the number of simple Allen relations in the disjunction (*i.e.*: for $r = \{b, o\}$, $|r| = 2$).

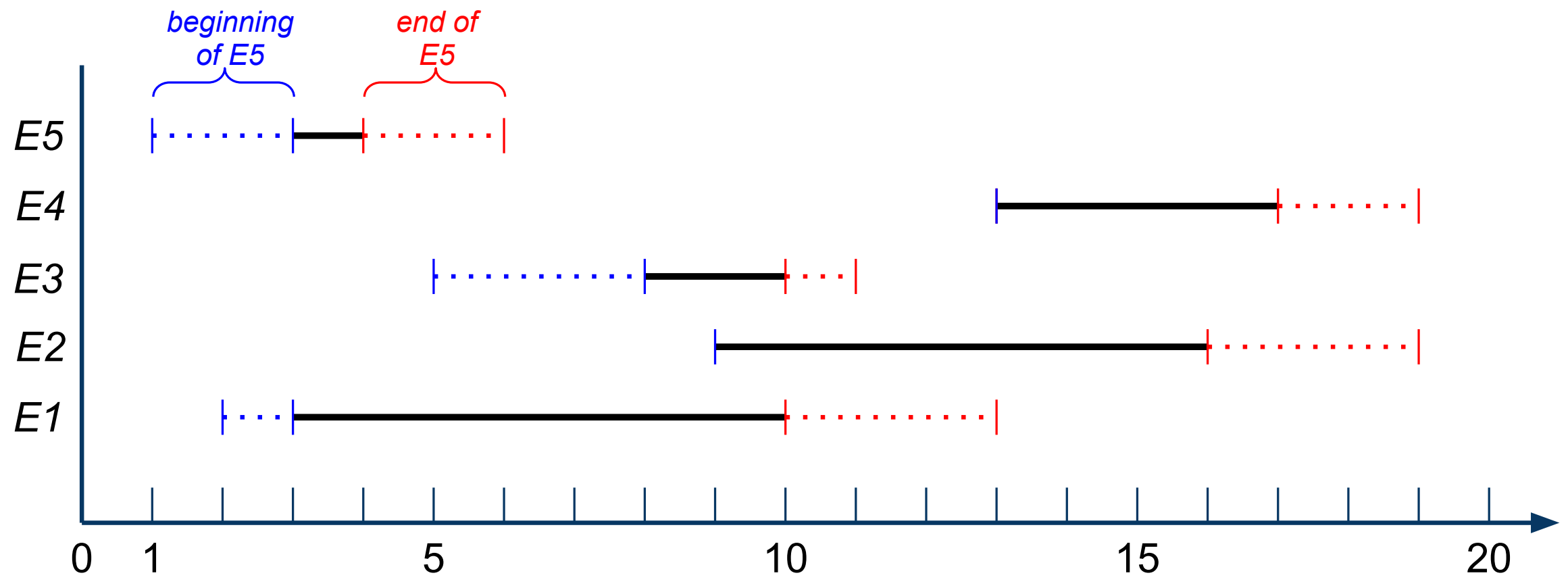

Figure 19: Example of random artificial graph with $N = 20$ (X-axis), $E = 5$ and $I = 3$ (maximum uncertainty for each endpoint). The indeterminacy leads to disjunctive relations, for example $((E5 < E3) \vee (E5 \text{ meets } E3) \vee (E5 \text{ overlaps } E3))$. When building graphs from this representation, relations are disjunctive but the graph is fully connected, which is not the case with real data like TimeBank.

Figure 20 shows the different values of strict recall and our temporal recall (called *point recall*) according to the proportion of relations kept in the graph, with lines joining values derived from the same graphs.

It is interesting to note that curves based on the minimal graphs are almost linear, while the simpler measures decrease slowly at first then sharply, in a more parabolic way. We analyze below (Section 6.2.2) the difference between being above or below "y=x".

This shows that the overall effect is observable roughly on every graph considered. The effect is even more marked when the number of events is higher.

We can also compare our point-based measure to the relaxed recall measure, similar to the measure used in the TempEval campaign and which computes an overlap between all simple and disjunctive relations on every edge of the graph. We also want to see the behavior of a measure based on the "core" set of relations extracted from the annotation as in our previous work (Tannier & Muller, 2008).

We can see in Figure 21 that the only measure that behaves linearly is the point-based one. Here, we have plotted the estimates of a parabolic regression on each data set, for clarity. The others measures show clearly parabolic behaviors, with different undesirable effects: relaxed recall is much too permissive when no information is actually provided (since the disjunction of all relations, which does not carry any information, gets a non-zero score), recall with respect to core relations is similar to strict recall, and is surprisingly even less linear. This effect is again stronger when the number of events increases.

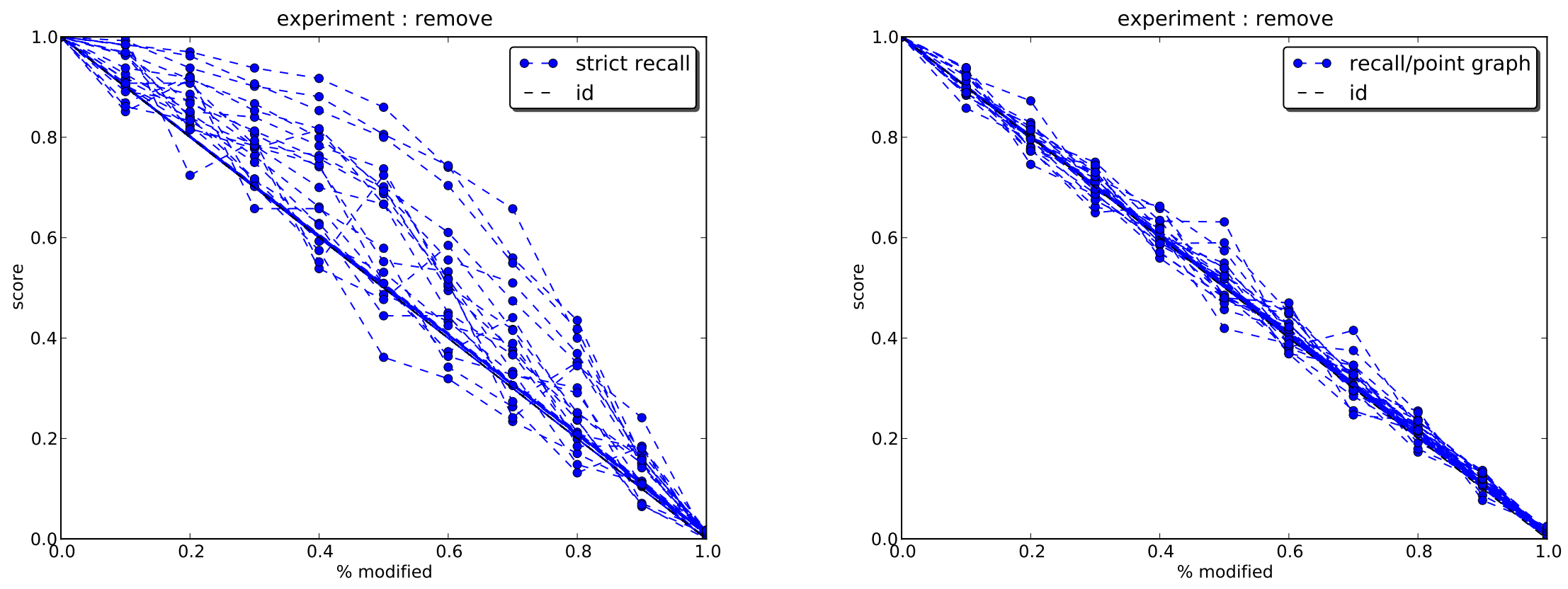

Figure 20: Behaviors for each graph with 30 events, strict recall (left) and point-based recall (right).

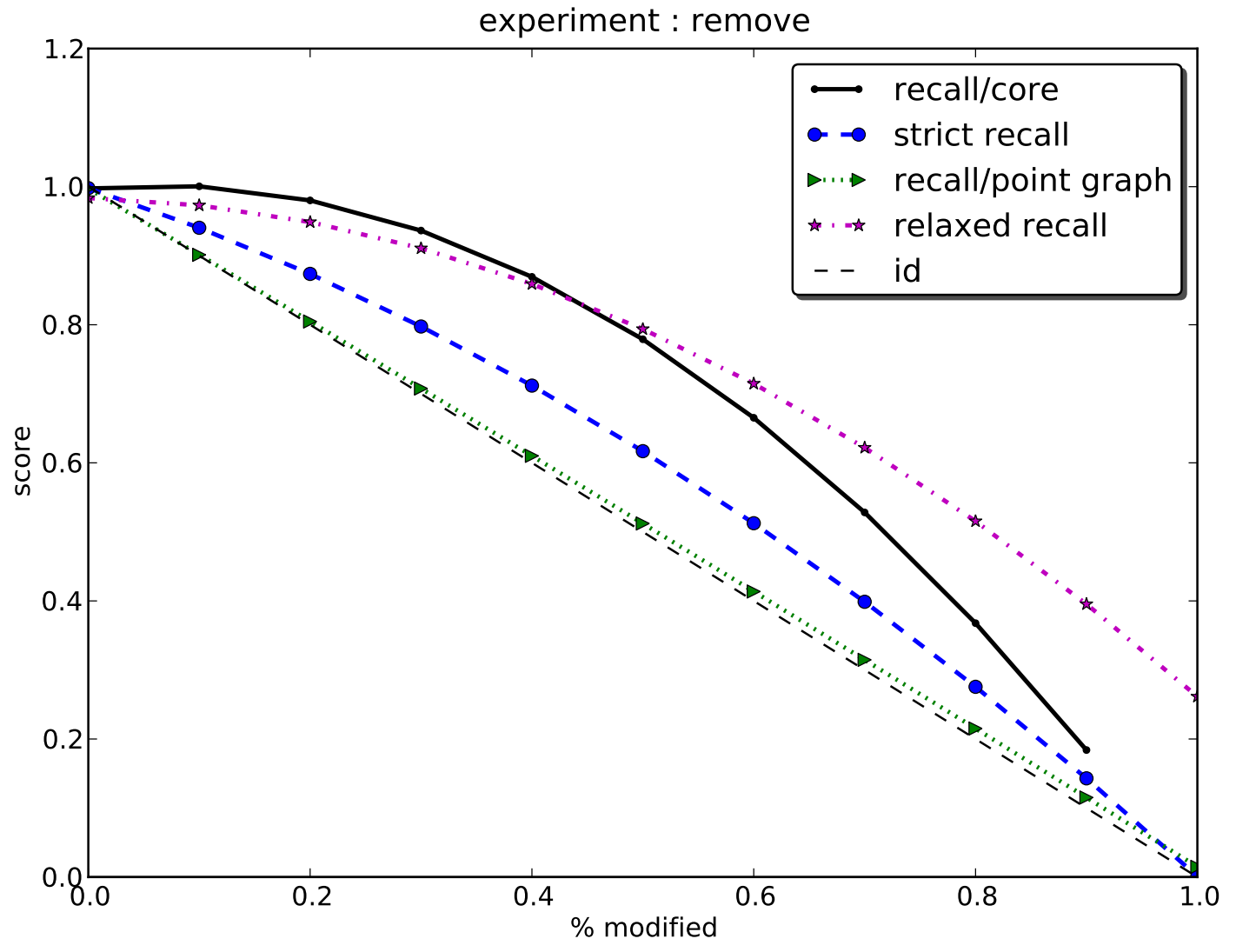

Figure 21: Parabolic regressions for all recall measures considered (30 event graphs).

Finally, Figure 22 shows the influence of different levels of "vagueness" of the annotation (and thus the underlying temporal description) on the measures used. We plotted the same experiment as above (recall with respect to the quantity of information removed) with the vagueness of the considered graphs as a third parameter. The surface shown is the departure from the "y=x" line, for readability's sake.

Here we can observe that less vague temporal descriptions have values mainly above the "y=x" line for "parabolic" measures, while vaguer data show a less obvious parabolic behavior, sometimes *below* that line. The curves for point-based recall seems insensitive to that aspect, which is consistent with the average behavior we already discussed.

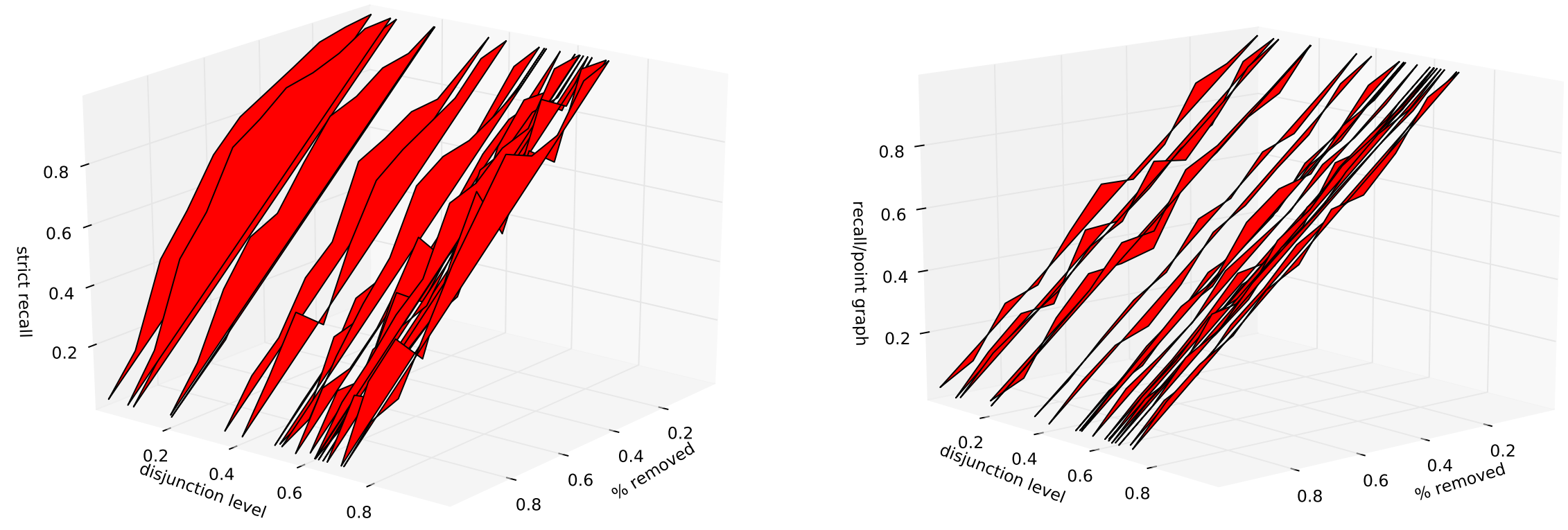

Figure 22: Influence of the vagueness of the annotation on the behavior of different measures (30 event graphs), with strict recall on the left and point recall on the right.

## 6.2 TimeBank Corpus

Before presenting the results of our evaluations on the real annotated data of the TimeBank corpus, we show some characteristics of the temporal graphs induced by the annotations. This helps to understand some of the differences between the behavior we observe on the synthetic temporal graphs and the more ecological ones.

### 6.2.1 ANALYSIS OF THE TIMEBANK CORPUS

The TimeBank corpus consists of 186 news report document (for about 65,000 tokens). It is made from news articles from the Associated Press, the Wall Street Journal, the L.A. Times and the San José Mercury News, and transcripts from broadcast news by CNN, ABC, VOA. These documents were initially collected for the DUC and ACE evaluation campaigns. The corpus in version 1.1 is available from http://www.timeml.org/site/timebank/timebank.html.[8]

Documents are annotated using the TimeML standard for tagging events and states, dates, times, and durations, and their temporal relations as well as various aspectual modalities. These entities are also tagged with some attributes: tense, aspect for verbs, values for dates and durations, normalized according to the ISO standard 8601.

Eventualities can be denoted by verbs, nouns, adjectives and some prepositional phrases. The temporal relations encode topological information between the time intervals of occurring eventualities, using relations that are equivalent to Allen relations, although they have different names. The TimeML standard specifies the relations between events, the anchoring of events to "times" (dates, hours) and the relations between times. There are about 7000 annotated relations between temporal entities. Additions to this set have been proposed (Bethard, Martin, & Klingenstein, 2007), in order to complete the annotations. As was noted by Setzer (2001), tagging temporal relations is hard for

8. A somewhat cleaned-up version, 1.2, is available via the Linguistic Data Consortium. We used the freely available version, while checking that the most recent version does not exhibit significant differences in our experiments.

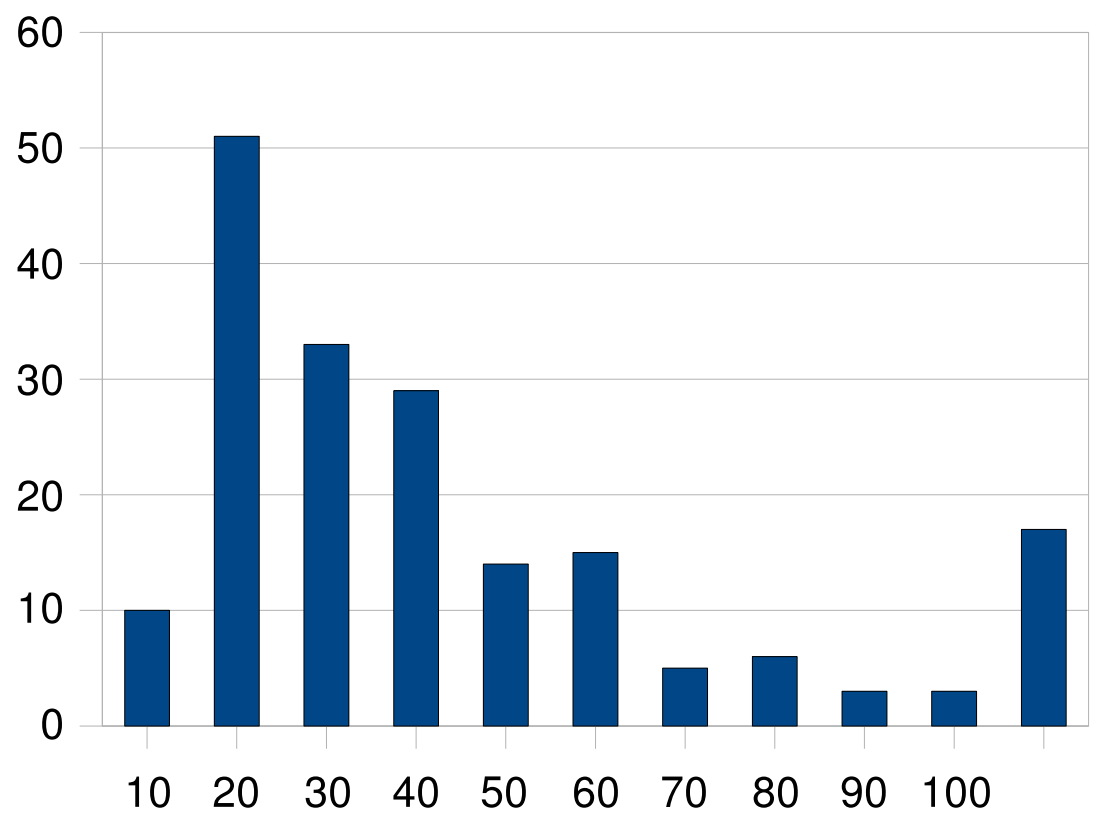

Figure 23: Frequencies of TimeBank texts with respect to their number of temporal entities.

| Mode | Raw | +time-time relations | Saturated | Saturated + t-t |
|---|---|---|---|---|
| Consistent annotations | 186 | 186 | 131 | 146 |
| Average number of relations | 43 | 58 | 134 | 281 |
| Average nb of components/text | 5.28 | 7.15 | 5.28 | 7.15 |
| Component Average size | 7.94 | 5.89 | 7.94 | 5.89 |
| Max component average size | 19.64 | 27.04 | 19.64 | 27.04 |

Table 4: Timebank1.1 statistics (Average number of temporal entities=42)

the annotators, who tend to miss relations, or produce inconsistent annotations. This is all the more obvious as the sizes of the texts increase, since then the number of possible links grows in the square of the number of temporal entities. Figure 23 shows the distribution of events among the texts. While annotators could be expected to keep track of all possible relations between events in small texts, the task was probably different for the numerous larger texts, and it has been noted that they tend to produce sets of disconnected temporal subgraphs on the majority of the corpus (Chambers & Jurafsky, 2008b).

Table 4 shows two important consequences when considering relations inferred from an annotation: (a) a lot of annotations in TimeBank 1.1 are actually inconsistent when the temporal graph is saturated using procedures introduced earlier in the paper and (b) each text gives rise to several connected components of various sizes. It is worse if we compute some missing (but obvious) relations between dates whose values are fully specified (noted t-t in the table). As we have seen, the procedure for checking consistency can only say that a set of relations is globally inconsistent, without a hint of which relation(s) should be isolated to repair that situation. These texts have thus been ignored in our evaluations.

The fact that the temporal graphs of some texts are actually split between components of different sizes has consequences when considering the size of the referent graph: a fully connected graph of $n$ entities can have up to $n \times (n-1)/2$ non-vague edges after saturation, while a text with scattered annotations might yield no other relation and a much smaller graph.

We have compared the number of relations present in a minimal graph obtained by transitive reduction with the number in the temporal closure of the interval-based graph, with respect to the number of events present in the text, on the whole TimeBank Corpus 1.1. (Figure 24). On this Figure, each point corresponds to a text, with its number of events along the x-axis and the number of relations in the reference used for evaluation with a given measure along the y-axis. It appears that minimal graph grows roughly linearly as expected (the variance being due to variable multiple branching when there is a lot of uncertainty). On the other hand, the temporal closure of the annotation is larger, much more irregular, with greater variance when the number of events grows. This is even worse for the "relaxed" measures which necessarily consider every possible edge between two events, even though it might only bear non-informative vague relations consisting of the disjunction of many relations. Note however that the reference size is not quadratic in the number of events, due to the high number of small self-connected components, as noted in the previous paragraph.

This has to be taken into account when considering an evaluation of a an entire corpus, and deciding what the contribution should be made by one text, or by a set of temporal relations. In the case of strict recall, the practice has been to add all edges from the temporal closures, possibly giving a given text a weight proportional to the square of its number of events. Micro-averaging the results on each text is probably not desirable either, giving too much importance to small texts with few relations. Having a reference with a size linear in the number of events solves the problem, providing a sort of smoothing with respect to that factor.

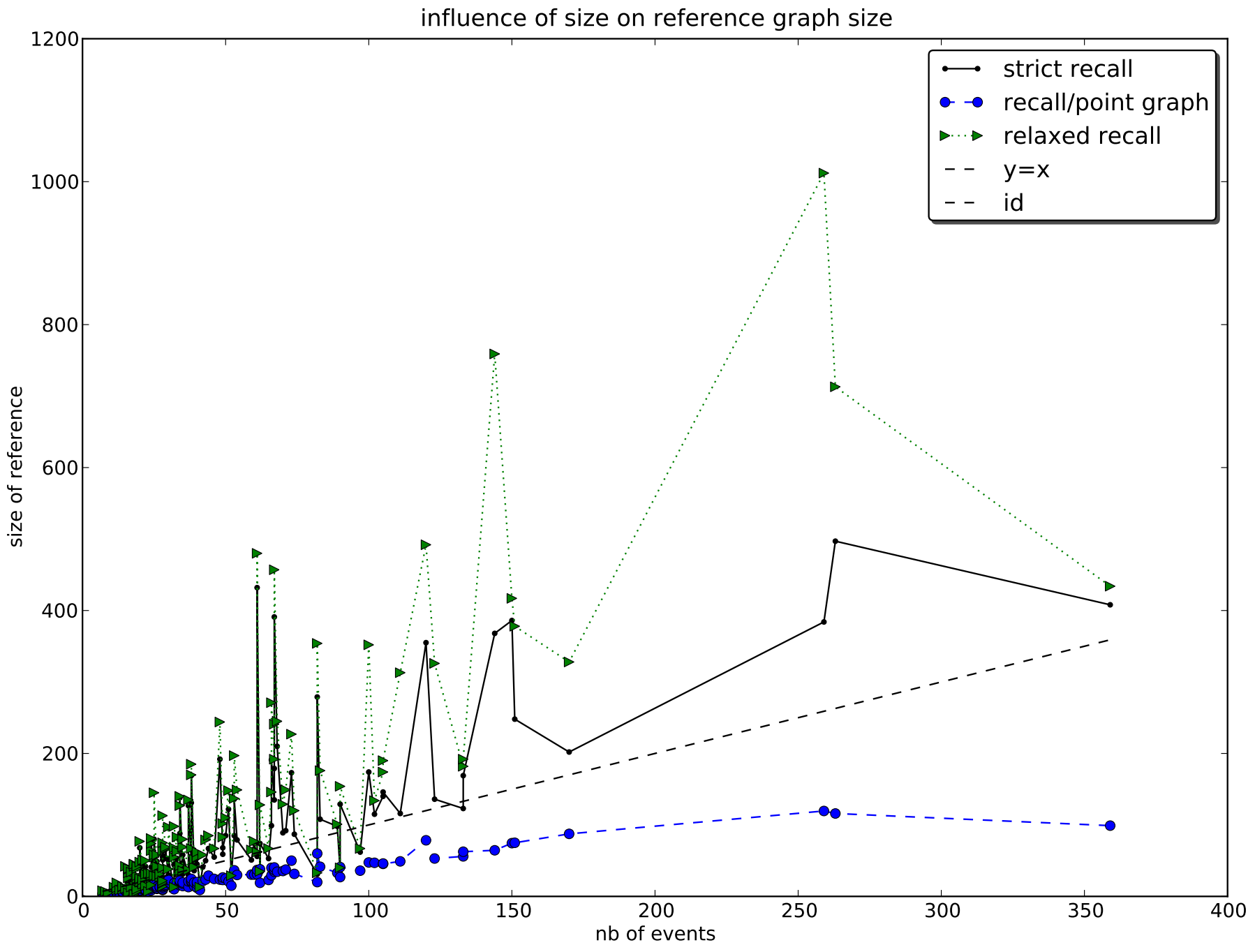

Figure 24: Number of relations considered in the reference *vs.* number of events in the texts, according to the measure used.

### 6.2.2 MEASURES ON TIMEBANK

**Recall** We performed the same experiments on TimeBank as we did on the synthetic data, and we can observe the same phenomenon (a linear decrease of point recall), while raw results are much more irregular (Figure 25), with a larger unstability when almost all relations are removed.

Note that while synthetic graph measures had values mainly above the “y=x” line, with some variation, annotated data show the same kind of parabolic behavior *below* that line.

The main difference here is how the reference is structured. The way generated graphs are built ensures that they are fully connected, and all relations from the saturated graphs are considered as the reference, since we have no reason to distinguish any subset in particular. On the other hand, in the human-annotated corpus, we only removed relations from the set of initial relations tagged by the annotators.

In the first case (artificial data), the reference graph is more resistant to the removal of random relations, since there is more redundancy, but in an extreme manner: it is only when enough relations are removed that some information is actually lost. This leads to a parabolic curve *above* the “y=x” line, showing that redundancy is improperly assessed by regular recall. The point graph is immune to this effect since the relevant relations are isolated first.

On the natural graphs on the contrary, the annotation is very “fragile”: annotators tend not to put too much redundant information in their annotation. Then, removing some of them removes also a lot of inferred relations at the same time (in quadratic quantity). This would not be a problem if the evaluation could stick to the annotated relations and if everyone tagged the same event pairs. But, as we already noticed, it is not the case. If we consider a system trying to build a temporal graph on a text from scratch, there is no reason to provide it with the same event pairs as the reference. For this reason, as soon as some redundancy is broken by removing relations, the recall improperly falls faster than “y=x” does. Again, the point-based graphs isolate the likely underlying models and their behavior is thus more controlled.

Remember also that the level of vagueness influences the shape of the curves, and that human annotations are less specified and thus highly disjunctive.

We can also assume than human choices, when annotating, are important. Annotated relations are probably regarded as central by the annotators, hence close to an ideal “core” set, especially since annotators tend to minimize the number of relations they tag. This assumption should be verified by more experiments on human assessments.

**Precision** To estimate the behavior of precision measures, we slightly changed the above experiment by switching more and more relations to different ones, thus “disturbing” the initial graph, while trying to keep it consistent. Again, we did this a number of times, and averaged the results on points with similar rates of undisturbed relations. The result, shown Figure 26, is restricted to smaller values of change, since as we increase the proportion of changed relations, the graph gets closer to a purely random annotation. So we arbitrarily kept values between 0 and 40% of relations changed at random. Of course, without any more control on the kind of change allowed, we generate a lot of different types of modifications, some minor as well as some which totally change the inferences drawn from the annotation. This is reflected in the huge variance of the results we observed on the detailed data (not shown here), even close to the origin, i.e. the unchanged graph. Still, the experiment seems to confirm that the point-based measure follows more closely the ideal “y=x” function, so it would be more stable than the others, with the previous caveat about the variance.

**System prediction** Finally, as a last indication of the importance of the evaluation methodology, we made a comparison of the behavior of the measures on the predictions of a real system. This is a rather tricky issue, since we want to see which measure distinguishes the better method from others, while the only way we have of evaluating the better method is the measure we want to assess. This was the motivation for the set of experiments we presented above. Still, it is legitimate to wonder if we can at least observe some differences in the context of a real system. One must then be cautious about the conclusions that can be drawn from this. In order to do so, we took an implementation of a temporal relation classifier reproducing standard work, as for instance in the work by Mani et al. (2006), that was used by one of the authors in another study (Denis & Muller, 2010). We applied it to TimeBank 1.1, and checked the correlations between the usual (strict) precision/recall and the counterparts based on transitive reduction. Figure 27 shows the relations between the scores on each text, with a subfigure for precision and one for recall. The spearman correlation, which estimates if one variable is monotonically related to the other without any other assumption, is 0.86 for precision, and 0.61 for recall, with very high significance levels ($p < 10^{-20}$). The tentative interpretation we can make is that while they are obviously related, both types of measures show a lot of variance on the different texts, and are obviously sensitive to differences in the predictions. Texts are indeed ordered very differently by the different sets of measures. This is especially true for recall. We also note that point-based scores are generally higher, which can be easily explained since relations on an event are correct only when relations about endpoints are correct at the same time. Inferences can still muddy the waters and yield different results in that respect, so it could be interesting to investigate the differences in more detail.

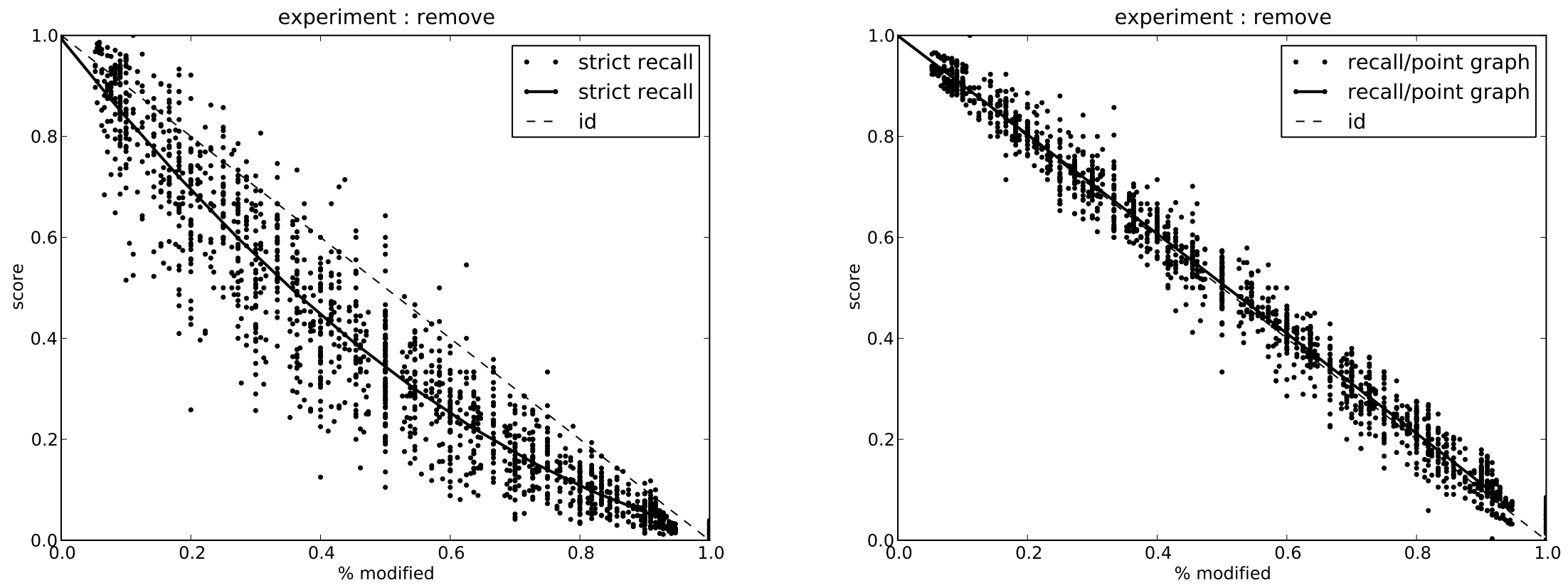

Figure 25: Behavior of recall measures on TimeBank according to the amount of temporal information removed, (left) strict recall and (right) recall on the point-based graph, with solid lines showing the parabolic regression.

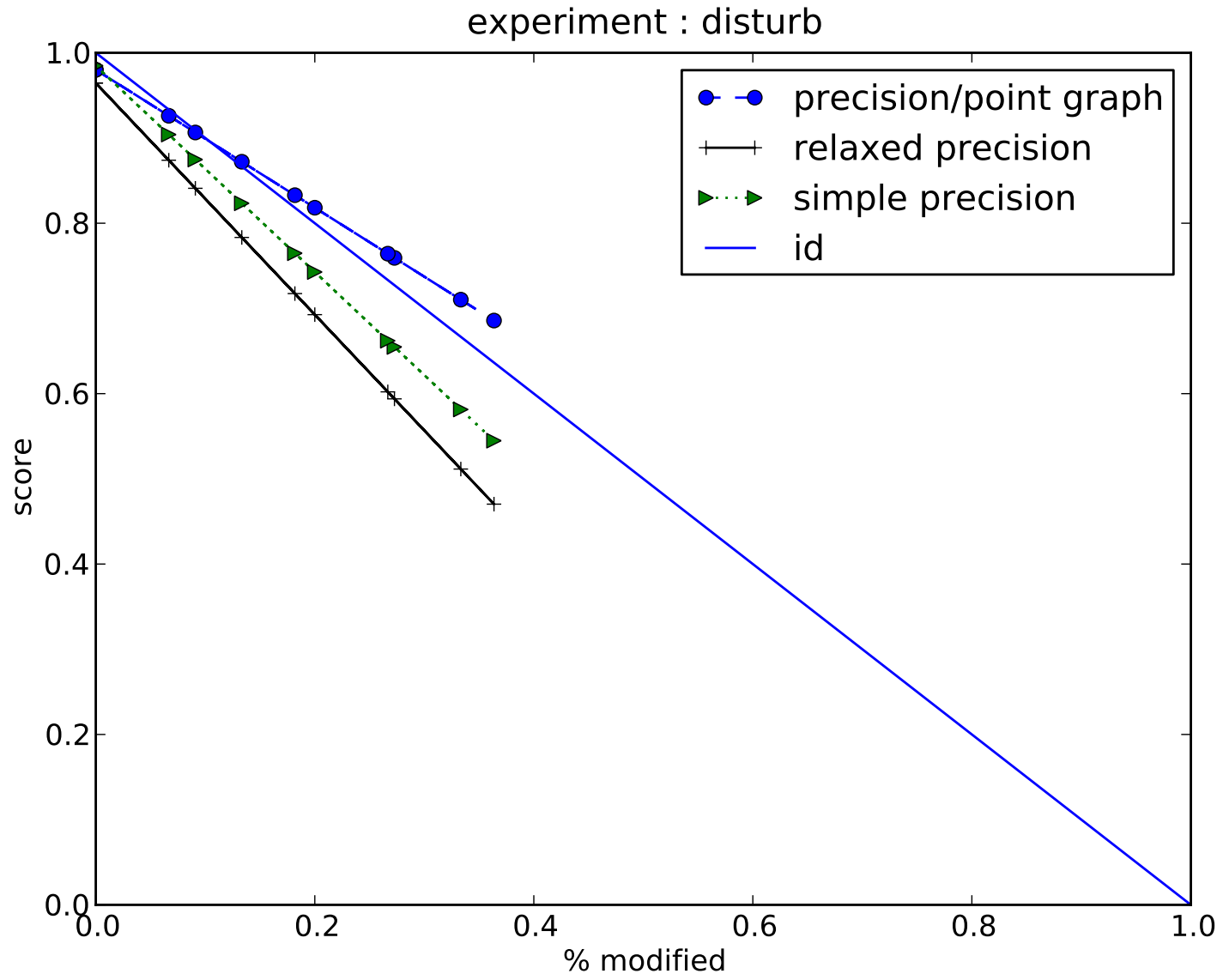

Figure 26: Linear regression of precision measures according to the amount of temporal information disturbed from the reference (TimeBank) in the range 0-40%

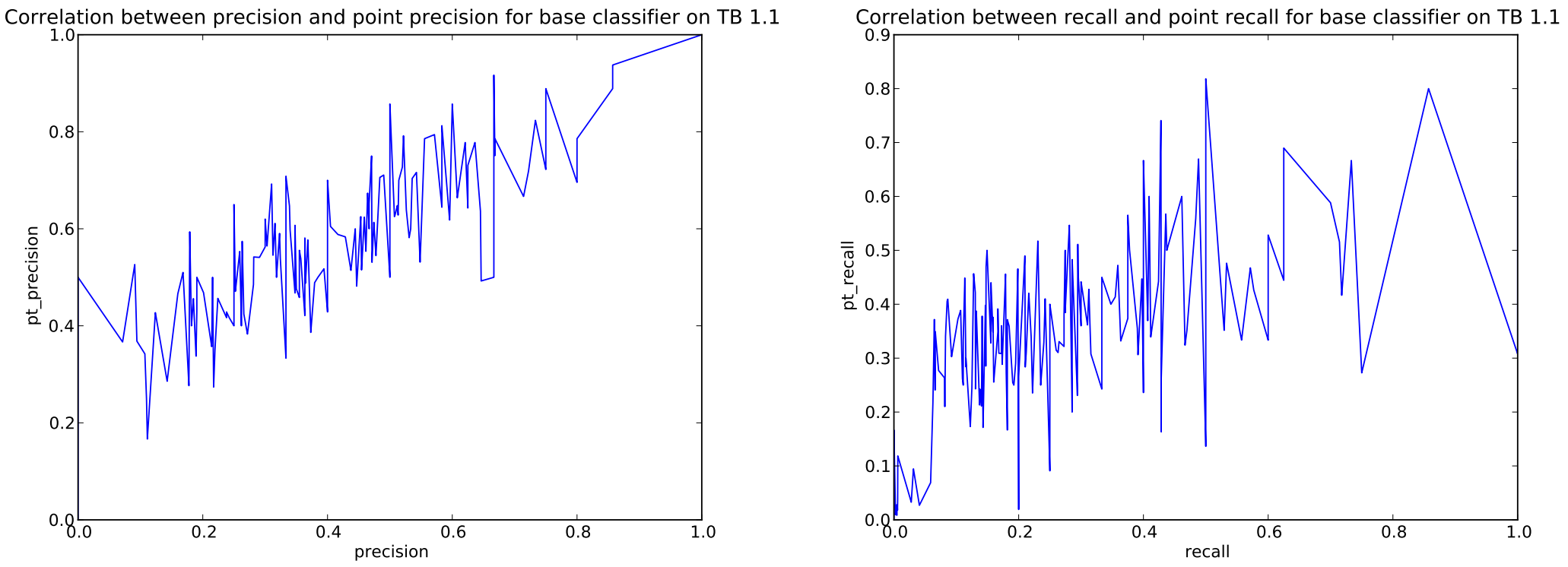

Figure 27: Correlations between classical measures and our proposal. Precisions are on the left, recalls on the right.

## 7. Conclusion

Comparing temporal constraints graphs is crucial in the task of extracting temporal information from texts, both from an evaluation point of view and in the perspective of incorporating global constraints in statistical learning procedures.

We argue here for comparison measures devoid of some of the biases inherent in the commonly used comparisons of closures of Allen-based temporal graphs. The measure is defined on the transitive reductions of the graph of (partially) ordered interval endpoints. Transitive reduction is conceptually intuitive, easy to compute and is unique in the cases considered. We have shown empirically that the behavior of this kind of measure is appropriate with the goals we had in mind.

We do not claim that ordering interval endpoints should be considered as the annotation provided by humans, only that the translation is possible and useful. It remains unclear if this could also be an acceptable way of presenting temporal information to humans, or how the resulting minimal graphs could be meaningfully re-translated into interval-based relations. We do not claim either that point relations should be the target of automated procedures to extract temporal information. The bulk of the work done on event relation classification deals primarily with extended intervals, as does the literature on temporal semantics (Steedman, 1997; Kamp & Reyle, 1993).

We plan to check our assumption that the procedure of translating interval constraints into endpoint constraints could be useful in the task of learning temporal constraints by integration of global constraints (for instance as a good indication of how close two temporal situations may be). It can also be useful when designing distances between structures in order to make structured predictions, in a manner similar to what is done on tree kernels for syntactic parsing (Collins & Duffy, 2002).

So far automated attempts only aim at predicting relations on given event-pairs, and most of them rely on a local learning strategy that does not take into account temporal constraints on the whole text. The few exceptions (Chambers & Jurafsky, 2008b; Bramsen et al., 2006) make use of subsets of relations, for instance a subset {before, after}, and any other relation is considered as "vague". Then they are able to use integer linear programming using transitivity constraints on the temporal order. This simplicity could be preserved in an endpoint translation of the full algebra.

The issues presented here are relevant also for other graph-based representations in natural language processing tasks, as long as there are inference issues: for instance discourse representations are often build as a set of rhetorical relations between segments (Marcu & Echihabi, 2002). Inference properties of such relations remain to be investigated, and automated processes are still quite rare (but see works by Sagae, 2009; Wellner, Pustejovsky, Havasi, Rumshisky, & Saurí, 2006; Subba & Di Eugenio, 2009), however some widely adopted structures behave like partial orders (narrative chains, topic elaborations) and thus should follow some of the patterns we have investigated. In case of partial annotations or partial agreement, finding a minimally equivalent representation could be used for comparison.

The limitations of this methodological study are also the limitations of our knowledge of the way human readers process the temporal information, as a whole, that can be found in texts. To the best of our knowledge, the psycholinguistics literature has little interest in that specific question. Work exists on *local* temporal interpretation, i.e. the way temporal order between two events is determined, with respect to linguistic and extra-linguistic factors (Zwaan & Razdvansky, 2001). In a more global context, studies have focussed on the way a reader builds a situation corresponding to a text (Speer, Zacks, & Reynolds, 2007; Zwaan, 2008), with temporal, spatial, causal aspects, and it seems events are grouped in time-frames, or sets of events with causal relations, and comprehension shifts from one time-frame representation to another. If these approaches are right, readers build mental models that correspond to the situation for a given time-frame, and do not explicitly record relations between time-frames. This could explain the structure of a lot of annotations (a set of small connected components), but does not say much about the nature of the knowledge represented. It remains to be seen how investigations into the process of human comprehension could be beneficial to computationally oriented approaches of such a process.

## 8. Acknowledgments

This work has benefited from tremendously helpful feedback from several colleagues, mainly Pierre Zweigenbaum and Pascal Denis, and Nicholas Asher. Pascal Denis also contributed to the development of the temporal tools used in this work, and allowed us to use some of the results of the system he developed with Philippe Muller for temporal relation identification. We also thank Michel Gagnon, who suggested the comparison of an annotation with a degraded version of itself as an evaluation of a measure, a long time ago.